\pdfoutput=1
\documentclass[10pt,twocolumn,letterpaper]{article}
\usepackage[pagenumbers]{cvpr}

\usepackage{cuted} %

\usepackage{currfile} %

\usepackage{caption} %

\definecolor{ftgood}{HTML}{1E9467}
\definecolor{ftbad}{HTML}{D64545}

\usepackage{booktabs,multirow,pifont,amssymb,amsmath,colortbl,makecell,enumitem,wasysym,microtype,accsupp}
\definecolor{cvprblue}{rgb}{0.21,0.49,0.74}
\usepackage[pagebackref,breaklinks,colorlinks,allcolors=cvprblue]{hyperref}
\hypersetup{pdftitle={Out of Sight, Not Out of Clues: Can Video LLMs Guess from Partial Evidence?},pdfauthor={Luoxi Zhang, Jiawei Liu, Yanlin Luo, Peiyang He, Junlin Zhu, Chun To Choi, Yang Xiong, Jiang Wu, Xingyu Zeng, Min Yang, Shiwen Ni}}

\usepackage{tikz}
\usetikzlibrary{positioning,arrows.meta,calc,fit,backgrounds}
\usepackage[most]{tcolorbox}
\definecolor{cNavy}{HTML}{243B53}\definecolor{cInk}{HTML}{1F2A3C}\definecolor{cBody}{HTML}{F3F5F8}\definecolor{cLine}{HTML}{C3CBD6}
\definecolor{cAl}{HTML}{E3EEF9}\definecolor{cGl}{HTML}{FDEBD6}\definecolor{cUl}{HTML}{E9ECF0}\definecolor{cOKl}{HTML}{E1F3EA}\definecolor{cBADl}{HTML}{FBE6E6}
\newcommand{\stag}[1]{{\setlength{\fboxsep}{1.2pt}\colorbox{cUl}{\scriptsize\sffamily\bfseries\color{cInk}\strut #1}}}
\newcommand{\lvA}{{\setlength{\fboxsep}{1.2pt}\colorbox{cA}{\scriptsize\sffamily\bfseries\color{white}\strut Answerable}}}
\newcommand{\lvG}{{\setlength{\fboxsep}{1.2pt}\colorbox{cG}{\scriptsize\sffamily\bfseries\color{white}\strut Guessable}}}
\newcommand{\lvU}{{\setlength{\fboxsep}{1.2pt}\colorbox{cU}{\scriptsize\sffamily\bfseries\color{white}\strut Unanswerable}}}
\newcommand{\tyA}{{\setlength{\fboxsep}{1pt}\colorbox{cAl}{\tiny\sffamily\bfseries\color{cA}\strut answer}}}
\newcommand{\tyG}{{\setlength{\fboxsep}{1pt}\colorbox{cGl}{\tiny\sffamily\bfseries\color{cG}\strut guess}}}
\newcommand{\tyR}{{\setlength{\fboxsep}{1pt}\colorbox{cUl}{\tiny\sffamily\bfseries\color{cU}\strut refuse}}}

\newtcolorbox{promptbox}[1][]{enhanced,breakable,colback=cBody,colframe=cLine,boxrule=0.4pt,arc=1.5pt,left=4pt,right=4pt,top=3pt,bottom=3pt,fontupper=\small,title={#1},coltitle=cInk,colbacktitle=cUl,fonttitle=\sffamily\bfseries\scriptsize,boxsep=1pt}
\tikzset{flowbox/.style={draw=cNavy,rounded corners=2pt,line width=0.5pt,fill=white,align=center,font=\scriptsize,inner sep=3pt},
 headbox/.style={rounded corners=2pt,fill=cNavy,text=white,font=\scriptsize\sffamily\bfseries,inner sep=3pt,align=center},
 softbox/.style={rounded corners=2pt,fill=cBody,draw=cLine,line width=0.4pt,font=\scriptsize,inner sep=3pt,align=left},
 arr/.style={-{Stealth[length=4pt]},line width=0.6pt,draw=cNavy}}

\def\confName{arXiv}
\def\confYear{2026}

\newcommand{\name}{ClueProbe}
\definecolor{cA}{HTML}{2B6CB0}\definecolor{cG}{HTML}{DD7A1A}\definecolor{cU}{HTML}{4A5568}
\definecolor{cOK}{HTML}{1F9D6B}\definecolor{cBAD}{HTML}{D64545}
\DeclareRobustCommand{\yes}{\BeginAccSupp{ActualText=yes}\textcolor{cOK}{\ding{51}}\EndAccSupp{}}
\DeclareRobustCommand{\no}{\BeginAccSupp{ActualText=no}\textcolor{cBAD}{\ding{55}}\EndAccSupp{}}

\newcommand{\tA}{Answerable}
\newcommand{\tG}{Guessable}
\newcommand{\tU}{Unanswerable}

\newcommand{\para}[1]{\noindent\textbf{#1}}

\setlist[itemize]{leftmargin=*,nosep}
\setlist[enumerate]{leftmargin=*,nosep}

\title{Out of Sight, Not Out of Clues:\\ Can Video LLMs Guess from Partial Evidence?}
\author{
Luoxi Zhang$^{1,*}$\quad
Jiawei Liu$^{1,*}$\quad
Yanlin Luo$^{1,*}$\quad
Peiyang He$^{2}$\quad
Junlin Zhu$^{3}$\quad
Chun To Choi$^{4}$\\
Yang Xiong$^{5}$\quad
Jiang Wu$^{6}$\quad
Xingyu Zeng$^{1}$\quad
Min Yang$^{7}$\quad
Shiwen Ni$^{1,\dagger}$\\[0.2em]
{\small $^{1}$Shenzhen University of Advanced Technology\quad
$^{2}$City University of Hong Kong\quad
$^{3}$The Education University of Hong Kong}\\
{\small $^{4}$The Hong Kong Polytechnic University\quad
$^{5}$Hong Kong Metropolitan University}\\
{\small $^{6}$The Chinese University of Hong Kong\quad
$^{7}$Shenzhen Institutes of Advanced Technology}\\
{\small $^{*}$Equal contribution\quad $^{\dagger}$Corresponding author}
}

\newcommand{\rUG}{0.83}

\begin{document}
\maketitle
\begin{abstract}
In real-world applications like egocentric recording, visual accessibility, and surveillance review, visual evidence for Video Large Language Models (Video LLMs) is often only partially captured. A reliable model should match its response to the evidence: answering when sufficient, guessing from visible clues when partial, and abstaining when absent. Yet existing benchmarks enforce a binary answerable-or-unanswerable split, leaving partial evidence unmeasured and obscuring whether failures arise from missing evidence or misjudging sufficiency. We introduce \name, a diagnostic benchmark to (1) separate perceiving evidence from judging sufficiency, and (2) validate our diagnosis via targeted fine-tuning. \name\ comprises 1,721 human-annotated questions across 598 natural videos, labeled Answerable, Guessable, or Unanswerable, alongside controlled perturbations altering wording or visual input. Evaluating 10 open-source and proprietary Video LLMs reveals that failures stem from judgment rather than perception: models perceive the evidence yet lack graded sufficiency awareness, answering or refusing outright while rarely guessing from partial clues. Targeted fine-tuning shows that this reflects missing supervision rather than insufficient capacity: supervising the guess level lifts appropriate guessing from 1.6\% to 40.4\% without extra false refusals, whereas conventional refusal tuning rejects 92\% of guessable questions. Data and code will be released at \url{https://github.com/suat-llm/ClueProbe}.
\end{abstract}

\section{Introduction}
\label{sec:intro}

\begin{figure}[t]
\centering
\includegraphics[width=\linewidth]{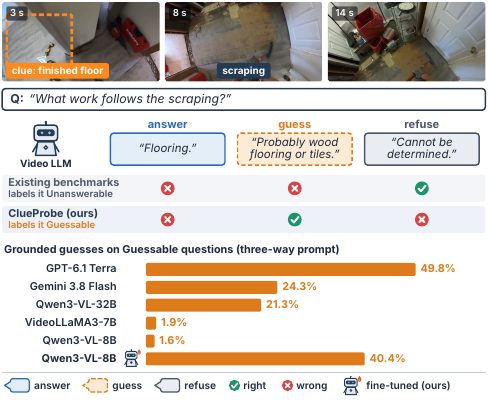}
\caption{\textbf{Out of sight, not out of clues.} The answer is not visible, but the finished floor next to the scraped patch narrows it. A benchmark with binary labels marks the question unanswerable and rewards the refusal, which discards the clue; \name\ labels it \tG\ and expects a grounded guess. Responses of Qwen3-VL-8B (abridged): the answer under the plain prompt, the refusal under the three-way prompt, and the guess after our guess-aware fine-tuning. Bottom: share of \tG\ questions answered with a grounded guess (G-Appr, \cref{tab:main}).}
\label{fig:teaser}
\end{figure}

Video large language models (Video LLMs) have advanced considerably on video question answering and captioning~\cite{bai2025qwen3vl,wang2025internvl35,zhang2024llavavideo,fu2025videomme,li2024mvbench}. As they are deployed on real recordings, such as egocentric assistance, visual accessibility, and surveillance review~\cite{grauman2022ego4d,gurari2018vizwiz,sultani2018surveillance}, they face questions whose evidence is only partially captured: the target may be occluded, too small to identify, or outside the recorded time span. A reliable model should then match its response to the evidence: answer when the video settles the question, make a grounded guess when visible clues narrow the answer, and abstain when nothing relevant is captured. Stating an unseen answer as fact misleads the user, whereas refusing whenever evidence is incomplete discards informative clues.

Current Video LLMs do not respond this way. In \cref{fig:teaser}, a worker scrapes a patch of floor next to finished flooring. Qwen3-VL-8B states what work follows as a fact and, once offered a guess as well, refuses instead, although the finished floor narrows the answer. Such failures can stem from distinct causes: the model may not see the evidence, or may see it but misjudge whether it suffices, and a misjudgment may reflect missing capacity or missing supervision. A single accuracy or refusal rate cannot tell these causes apart.

Existing benchmarks address only isolated facets of this problem (\cref{tab:compare}). Video answerability benchmarks label questions as answerable or unanswerable~\cite{yu2026virtuebench,yoon2025uvqa,zhong2026storm}, image abstention benchmarks perturb static inputs to test refusal~\cite{zhu2026mmooc,madhusudhan2026mmaqa,chandu2025certainly}, and refusal alignment trains models to decline unanswerable questions~\cite{yoon2025uvqa}. None measures partial evidence in video, separates seeing the evidence from judging its sufficiency, or tests whether supplying the missing supervision repairs the failure. Two research questions therefore remain open: 1) Under incomplete evidence, do Video LLMs fail to see the evidence, or fail to judge its sufficiency? 2) Does supervising the missing guess level repair this failure where conventional refusal tuning does not?

\begin{table}[t]
\centering
\footnotesize
\setlength{\tabcolsep}{3.2pt}
\resizebox{\linewidth}{!}{%
\begin{tabular}{@{}lcccccc@{}}
\toprule
 & & \multicolumn{2}{c}{Labels \& scoring} & \multicolumn{2}{c}{One input changed} & Fine-tuning \\
\cmidrule(lr){3-4}\cmidrule(lr){5-6}\cmidrule(l){7-7}
Benchmark & Video & \makecell{Guessable\\level} & \makecell{Guess\\response} & \makecell{Prompt or\\question} & \makecell{Visual\\input} & \makecell{Contrasting\\targets} \\
\midrule
VirtueBench~\cite{yu2026virtuebench} & \yes & \no & \no & \yes & \yes & \no \\
UVQA~\cite{yoon2025uvqa} & \yes & \no & \no & \no & \no & \no \\
STORM-Bench~\cite{zhong2026storm} & \yes & \no & \no & \no & \yes & \no \\
AbstainEQA~\cite{wu2026abstaineqa} & \yes & \no & \no & \yes & \no & \no \\
MMOOC~\cite{zhu2026mmooc} & \no & \no & \no & \yes & \no & \no \\
MM-AQA~\cite{madhusudhan2026mmaqa} & \no & \no & \no & \yes & \yes & \no \\
CertainlyUnc.~\cite{chandu2025certainly} & \no & \no & \no & \yes & \yes & \no \\
AQuA~\cite{jang2026aqua} & \no & \no & \yes & \no & \no & \no \\
\midrule
\textbf{\name\ (ours)} & \yes & \yes & \yes & \yes & \yes & \yes \\
\bottomrule
\end{tabular}}
\caption{\textbf{Comparison with related benchmarks} by evidence labels, response scoring, inputs changed on the same item, and fine-tuning targets.}
\label{tab:compare}
\end{table}

\begin{figure*}[t]
\centering
\includegraphics[width=\linewidth]{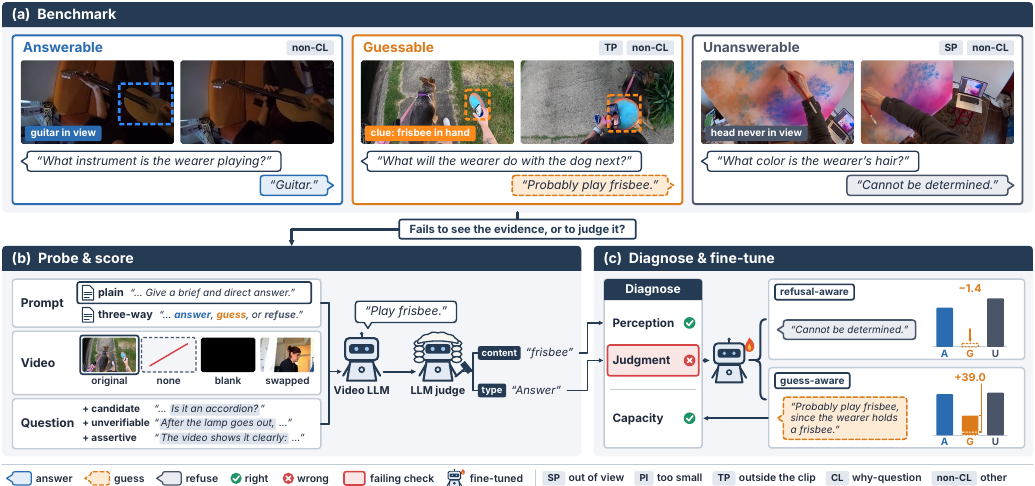}
\caption{\textbf{Overview of \name.} (a) Levels follow how much of the answer the video shows (dashed boxes: answer or clue); tags give the gap and the question logic. (b) Each question is asked under two prompts, a subset also with one input changed; an LLM judge scores content and type separately. Qwen3-VL-8B reaches the clue-supported answer but states it as fact. (c) Content feeds the perception check, type the judgment check (\cref{sec:metrics}), and fine-tuning the capacity check: tuned on targets that differ only for \tG\ questions, the language model learns to guess, so the failure lies in supervision. Bars: A-Acc, G-Appr, U-Ref under the three-way prompt (\cref{tab:main}), with the paired change in G-Appr (\cref{tab:ft}).}
\label{fig:overview}
\end{figure*}

To answer these questions, we introduce \name, a diagnostic benchmark for Video LLMs under incomplete evidence (\cref{fig:overview}). It contains 1,721 human-annotated questions on 598 natural videos from Charades~\cite{sigurdsson2016charades} and Ego4D~\cite{grauman2022ego4d}, each labeled \tA, \tG, or \tU\ by how much evidence the video provides; Guessable questions carry a reference guess and its visible clue, and incomplete evidence is tagged by its cause. \name\ scores the response type separately from the content and pairs questions with controlled perturbations that change only the question text or only the video.

Evaluating ten Video LLMs (two proprietary and eight open-source) shows that failures stem from judgment rather than perception. Models see the evidence: they answer most Answerable questions, reject wrong candidate answers when the video is present, and reach the clue-supported answer on most Guessable questions. Yet by default they state that answer as fact and almost never guess. Even the models that flag many blank or mismatched clips without being asked miss most natural gaps, where the video shows the scene but not the answer. A prompt that offers a guess switches refusal on but sends partial evidence to refusal as well, and refusal is rarest for temporal gaps, which only video can pose.

Fine-tuning supports this diagnosis. Changing only the training target of Guessable questions raises appropriate guessing from 1.6\% to 40.4\% without extra false refusals and transfers from Charades to Ego4D videos, whereas conventional refusal tuning on the same data raises false refusals even on Answerable questions.

By combining evidence-level diagnosis with targeted fine-tuning, \name\ offers a systematic framework for evaluating and improving how Video LLMs respond to incomplete evidence. Our main contributions are threefold:
\begin{itemize}
\item \textbf{Diagnostic Benchmark.} We introduce \name, which labels video questions by three levels of evidence sufficiency and pairs them with controlled perturbations of the question text and the video, to separate seeing the evidence from judging its sufficiency.
\item \textbf{Evidence-Level Diagnosis.} Evaluating ten Video LLMs shows that failures stem from judgment rather than perception: models lack a graded sense of sufficiency, answering by default, refusing when instructed, and rarely guessing from partial clues.
\item \textbf{Fine-Tuning Validation.} By contrasting guess-level supervision with conventional refusal tuning, we show that the diagnosis guides model improvement: guess-level supervision raises appropriate guessing without extra false refusals and transfers to unseen video sources.
\end{itemize}

\section{Related Work}
\label{sec:related}

\subsection{Video Question Answering Benchmarks}
Video LLMs are commonly evaluated on multiple-choice and open-ended video question answering, such as MVBench~\cite{li2024mvbench} and Video-MME~\cite{fu2025videomme}, which report accuracy on questions that the video answers. One line of work asks where an answer comes from: single-frame and blind baselines show that many questions can be solved from language priors~\cite{lei2023singleframe,chen2024rightway}, CG-Bench~\cite{chen2025cgbench} and CaST-Bench~\cite{zhang2026castbench} score whether answers are grounded in the right segments, and HERBench~\cite{benami2026herbench} measures how many frames an answer requires. Another line asks whether an answer exists at all: VirtueBench~\cite{yu2026virtuebench} removes frames from long videos, UVQA~\cite{yoon2025uvqa} generates questions beyond the content of a video, and STORM-Bench~\cite{zhong2026storm} queries a stream before the evidence appears. All of them treat a question as either answerable or not. \name\ adds the level in between, where the answer is hidden but narrowed by visible clues, and scores a hedged guess as the correct response to it.

\subsection{Abstention under Missing Visual Evidence}
Video LLMs describe content that is absent from the video~\cite{wang2024videohallucer,li2025vidhalluc}, and language models can be swayed by confident user statements~\cite{sharma2024sycophancy}. Whether a model should decline instead has been studied mostly on images, in protocols that video benchmarks inherit. Following SQuAD~2.0~\cite{rajpurkar2018squad2} and VizWiz~\cite{gurari2018vizwiz}, Reliable VQA~\cite{whitehead2022reliable}, UPD~\cite{miyai2025upd}, MoHoBench~\cite{zhu2026mohobench}, MM-AQA~\cite{madhusudhan2026mmaqa}, and MMOOC~\cite{zhu2026mmooc} test whether models decline image questions made unanswerable by false premises, missing context, or perturbed evidence; CertainlyUncertain~\cite{chandu2025certainly} separates epistemic from aleatoric uncertainty, AbstainEQA~\cite{wu2026abstaineqa} extends abstention to embodied agents, and AQuA~\cite{jang2026aqua} grades ambiguous questions by how clearly they identify their referent. On synthetic physics videos, TRAPSBench~\cite{pramono2026trapsbench} finds that models encode the need to abstain without expressing it. These studies score abstention as a binary choice, and those that change the visual input do so by editing or removing it~\cite{yu2026virtuebench,madhusudhan2026mmaqa,chandu2025certainly}. \name\ instead tests natural gaps, where the video shows the scene but not the answer, and on the same items changes either the question text or the video, which shows whether a response follows what the model sees or what it reads.

\subsection{Fine-Tuning for Abstention}
Refusal-aware tuning teaches language models to decline questions beyond their knowledge~\cite{zhang2024rtuning}, and related work trains them to verbalize uncertainty~\cite{lin2022teaching}. For Video LLMs, UVQA~\cite{yoon2025uvqa} aligns models to refuse unanswerable questions through supervised and preference-based tuning; on images, MoHoBench~\cite{zhu2026mohobench}, CertainlyUncertain~\cite{chandu2025certainly}, and MMOOC~\cite{zhu2026mmooc} report gains from similar post-training, and AQuA~\cite{jang2026aqua} trains strategy-aware responses to ambiguous questions. These methods train a chosen response for missing or ambiguous evidence and report the resulting gain. \name\ instead uses fine-tuning as a controlled test of a diagnosis: two runs share data and settings and differ only in the target for Guessable questions, which shows whether the missing level is a matter of supervision.

\section{\name\ Benchmark}
\label{sec:benchmark}

\subsection{Benchmark Overview}
\label{sec:design}
\name\ has two objectives: (1) to determine whether a Video LLM that mishandles incomplete evidence fails to see the evidence or fails to judge its sufficiency, and (2) to test whether a judgment failure can be repaired by supervision. Both require what binary video QA lacks: ground truth that separates partial from absent evidence, scores that separate what a model says from how it says it, and inputs that change one factor at a time (\cref{fig:overview}).

\para{Evidence levels.} Under binary labels, a hedged inference from a visible clue counts as an error just like a fabricated fact. \name\ instead labels each question by how much of its answer the video provides: \tA\ (the video settles it), \tG\ (it is not visible, but visible clues narrow it), or \tU\ (nothing visible bears on it). The expected responses are a definite answer, a guess that states its uncertainty (\eg \emph{``probably''}), and a refusal.

\para{Gap tags.} Questions with incomplete evidence also record why the answer is not visible, so that refusals can be traced to the kind of gap: \emph{spatial} (SP), out of frame or occluded, such as the wearer's own hair in a first-person clip; \emph{perceptual} (PI), in view but too small or blurred, such as the product name on a bucket label; or \emph{temporal} (TP), before or after the clip, such as what the wearer will do next with the dog, a gap that only video poses (\cref{fig:overview}). Causal questions (CL; \emph{``why \dots?''}) are flagged at every level, since they invite commonsense explanations beyond what the video shows. \cref{sec:judge} tests whether refusal follows these tags.

\para{Evaluation settings.} A model that judges sufficiency should change its response when the evidence changes and keep it when only the wording changes. \name\ therefore asks every question under two instructions and re-asks a subset with exactly one input changed:
\begin{itemize}
\item \textbf{Plain prompt (P).} Ask for a brief answer, as a user would. This tests the default response to incomplete evidence.
\item \textbf{Three-way prompt (T).} Permit a direct answer, a hedged guess, or a refusal. This tests whether the model grades the evidence once all three responses are offered.
\item \textbf{Video perturbations.} Remove the video, replace it with a blank clip, or swap in a clip from the other source. This tests whether the response depends on the evidence at all.
\item \textbf{Candidate in the question.} Embed a correct or a wrong candidate (\emph{``\dots Is it an accordion?''}). This tests whether suggested content is checked against the video.
\item \textbf{Unverifiable content.} Presuppose an unseen event (\emph{``After the lamp goes out, \dots''}) or ask about the time around the clip (\emph{``\dots put it back afterwards?''}). This tests whether the model accepts what the video cannot confirm.
\item \textbf{Assertive wording.} Claim that the video clearly shows the answer, or add up to four true but redundant qualifiers. This tests whether wording that adds no content shifts the response.
\end{itemize}

\subsection{Benchmark Construction and Quality Control}
\label{sec:build}
\para{Videos.} Natural recordings lose evidence through the gaps above rather than through synthetic corruption. We use silent clips from Charades~\cite{sigurdsson2016charades}, third-person recordings of indoor household activities, and Ego4D~\cite{grauman2022ego4d}, first-person recordings of daily tasks with frequent camera motion and partial views of the workspace. The two sources differ in viewpoint, so every finding is also checked within each source.

\para{Annotation.} Annotators write each question about a target that its wording identifies uniquely, without leaking the answer or presupposing unseen events (Supp.~\ref{supp:data}). The level is decided from the video alone, in order: \tA\ if the frames show the answer at the granularity asked; otherwise \tG\ if a visible object, state, or action that another viewer could verify narrows the answer to a few candidates; otherwise \tU. \tG\ questions record the reference guesses and their clue, and \tU\ questions the missing evidence.

\para{Quality control.} Every question, reference, clue, and tag was checked a second time against the frames, and flawed items were removed. Two annotators independently labeled 60 questions and agreed on 57 (95.0\%).

\para{Statistics.} The main set has 1,721 questions (661 \tA, 429 \tG, 631 \tU) on 598 clips, 299 from each source, which last 3 to 60 seconds (median 31) and carry up to three questions each. Of the questions with incomplete evidence, 392 have a spatial, 364 a perceptual, and 303 a temporal gap, and 258 questions across all levels are causal, nearly all from Ego4D (Supp.~\ref{supp:data}). The perturbation set changes one input of 300 base questions (200 \tA, 100 \tU) in 1,897 variants and adds one to four redundant qualifiers to 198 questions, for 6,794 queries per model.

\begin{table*}[t]
\centering
\small
\setlength{\tabcolsep}{3.0pt}
\resizebox{\textwidth}{!}{%
\begin{tabular}{@{}l cccccc ccccccc ccc cc c@{}}
\toprule
 & \multicolumn{6}{c}{Plain prompt} & \multicolumn{7}{c}{Three-way prompt} & \multicolumn{5}{c}{U-Ref by tag (three-way)} & No video \\
\cmidrule(lr){2-7}\cmidrule(lr){8-14}\cmidrule(lr){15-19}\cmidrule(l){20-20}
Model & A-Acc & G-Hit & G-Appr & U-Ref & Avg & ES & A-Acc & A-FR\,$\downarrow$ & G-Appr & G-FR\,$\downarrow$ & U-Ref & Avg & ES & SP & PI & TP & CL & non-CL & U-Ref \\
\midrule
\multicolumn{20}{@{}l}{\textit{Proprietary}}\\
GPT-6.1 Terra~\cite{openai2026gpt61terra} & \textbf{97.7} & \textbf{85.5} & 3.3 & 47.9 & 49.6 & 18.2 & \underline{92.4} & \underline{0.3} & \underline{49.8} & 19.6 & 78.0 & \textbf{73.4} & \underline{44.9} & 84.1 & 65.3 & 85.3 & 66.7 & 78.0 & 5.2 \\
Gemini 3.8 Flash~\cite{google2026gemini38flash} & \underline{96.7} & \underline{81.8} & 2.1 & 39.4 & 46.1 & 14.1 & \textbf{95.7} & 0.5 & 24.3 & 34.8 & 86.2 & 68.7 & 40.9 & \underline{91.9} & 75.5 & \underline{91.8} & \underline{80.8} & 83.8 & 15.8 \\
\midrule
\multicolumn{20}{@{}l}{\textit{Open-source}}\\
Qwen3-VL-32B~\cite{bai2025qwen3vl} & 94.4 & 80.6 & 1.6 & 23.4 & 39.8 & 8.2 & 90.8 & 0.9 & 21.3 & 23.8 & 77.8 & 63.3 & 33.4 & 79.8 & 74.1 & 79.8 & 36.0 & 81.3 & 7.1 \\
Qwen3-VL-8B-Thinking~\cite{bai2025qwen3vl} & 88.8 & 79.4 & 1.2 & 26.5 & 38.8 & 9.0 & 87.4 & 0.8 & 28.8 & 28.7 & 75.8 & 64.0 & 34.5 & 78.4 & 81.3 & 64.6 & 27.0 & 80.0 & \textbf{42.6} \\
Qwen3-VL-8B~\cite{bai2025qwen3vl} & 88.6 & 79.2 & 0.0 & 10.6 & 33.1 & 3.9 & 89.3 & 2.7 & 1.6 & 51.3 & 86.8 & 59.2 & 31.1 & 90.5 & 89.8 & 77.3 & 70.7 & 89.0 & 9.6 \\
Qwen3-VL-2B~\cite{bai2025qwen3vl} & 83.4 & 68.5 & 1.6 & 6.1 & 30.4 & 2.4 & 81.2 & 0.5 & 12.9 & \underline{10.3} & 39.2 & 44.4 & 12.9 & 35.9 & 59.1 & 17.9 & 10.7 & 40.0 & 6.2 \\
InternVL3.5-8B~\cite{wang2025internvl35} & 87.9 & 68.8 & 0.0 & 5.0 & 31.0 & 1.6 & 85.3 & \textbf{0.2} & 0.9 & \textbf{3.3} & 32.6 & 39.6 & 10.8 & 24.7 & 56.0 & 13.6 & 1.4 & 36.6 & \underline{23.3} \\
InternVideo2.5-8B~\cite{wang2025internvideo25} & 87.7 & 76.1 & 0.0 & 2.9 & 30.2 & 1.0 & 87.3 & 0.9 & 2.6 & 19.5 & 63.7 & 51.2 & 22.2 & 67.7 & 74.5 & 42.9 & 18.9 & 67.9 & 0.2 \\
VideoLLaMA3-7B~\cite{zhang2025videollama3} & 88.3 & 76.1 & 0.0 & 3.2 & 30.5 & 1.2 & 81.2 & 7.6 & 1.9 & 59.7 & \underline{87.8} & 57.0 & 26.9 & 86.9 & \underline{92.6} & 82.8 & 75.7 & 90.2 & 4.6 \\
LLaVA-Video-7B~\cite{zhang2024llavavideo} & 86.2 & 72.0 & 0.7 & 5.1 & 30.7 & 1.9 & 74.7 & 9.8 & 20.4 & 31.1 & 73.6 & 56.2 & 20.3 & 75.8 & 82.9 & 57.8 & 16.0 & 76.6 & 0.3 \\
\midrule
\multicolumn{20}{@{}l}{\textit{Qwen3-VL-8B after fine-tuning$^{\dagger}$}}\\
+ Refusal-aware & 92.1 & 51.9 & 2.1 & \underline{74.2} & \underline{56.1} & \underline{25.9} & 79.9 & 15.7 & 0.0 & 91.6 & \textbf{99.2} & 59.7 & 36.1 & \textbf{98.8} & \textbf{99.5} & \textbf{99.4} & \textbf{96.0} & \textbf{100.0} & -- \\
+ Guess-aware (ours) & 93.2 & 76.2 & \underline{11.7} & 47.4 & 50.8 & 22.5 & 84.4 & 2.1 & 40.4 & 30.8 & 87.3 & 70.7 & 43.9 & 88.5 & 87.0 & 85.9 & 41.3 & \underline{93.9} & -- \\
+ Guess-aware, both prompts (ours) & 85.9 & 54.4 & \textbf{37.9} & \textbf{88.7} & \textbf{70.8} & \textbf{46.1} & 88.0 & 1.1 & \textbf{50.2} & 17.5 & 80.2 & \underline{72.8} & \textbf{48.5} & 79.8 & 77.3 & 84.7 & 42.7 & 86.2 & -- \\
\bottomrule
\end{tabular}}
\caption{\textbf{Main results (\%).} Metrics are defined in \cref{sec:metrics}; ES is at most 61.6 for a model that only answers or refuses. The tag columns split U-Ref under the three-way prompt by gap and by question logic (CL and non-CL within Ego4D). \emph{No video}: plain prompt without the video. $^{\dagger}$Fine-tuned in \cref{sec:finetune}; rates count invalid outputs as failures. Higher is better except for $\downarrow$. \textbf{Bold}: best; \underline{underlined}: second best.}
\label{tab:main}
\end{table*}

\subsection{Metrics and Diagnostic Criteria}
\label{sec:metrics}
\para{Decoupled scoring.} A single accuracy conflates what a model knows with how it states it. An LLM judge therefore first assigns each response a type: a definite \emph{answer}, a \emph{guess} that states its uncertainty, or a \emph{refusal} stating that the video cannot determine the answer. For answers and guesses it then checks the content against the reference answer or, for \tG\ questions, whether it falls in the class of the reference guess and agrees with its clue. Against human labels, the judge reaches $\kappa=0.96$ on type and $0.84$ on content (Supp.~\ref{supp:prompts}).

\para{Level metrics.} A-Acc, G-Appr, and U-Ref are the shares of \tA, \tG, and \tU\ questions that receive the expected response: a correct answer, a correct guess, and a refusal, respectively; Avg is their mean. A-FR and G-FR are the shares of \tA\ and \tG\ questions that are refused. G-Hit is the share of \tG\ questions with correct content, whether stated as an answer or as a guess, so G-Hit minus G-Appr is the share on which the clue-supported answer is reached but stated as fact.

\para{Evidence sensitivity.} A reliable response type also tells the user how much evidence lies behind the content. We measure how much the type $R$ reveals about the level $L$ by normalized mutual information,
\begin{equation}
\text{ES} = 100\cdot I(L;R)\,/\,H(L),
\label{eq:es}
\end{equation}
which is 100 when the type identifies the level and 0 when it is the same at every level. A model that only answers or refuses scores at most 61.6 on \name\ (Supp.~\ref{supp:es}); higher values require guessing.

\para{Diagnostic criteria.} Failures of perception and of judgment call for different remedies, so the checks are applied in order:
\begin{enumerate}
\item \textbf{Perception.} A model sees the evidence if removing the video lowers A-Acc, if with the video it accepts correct candidates far more often than wrong ones, and if its G-Hit exceeds the value without the video. Otherwise it is limited by perception.
\item \textbf{Sufficiency judgment.} A model that sees the evidence judges its sufficiency if its response type tracks the level (high ES), changes when the video is removed, blank, or swapped, and does not take up content from the question that the video cannot confirm. Otherwise it is limited by judgment.
\item \textbf{Capacity or supervision.} A judgment failure reflects missing supervision rather than missing capacity if the model detects missing evidence when instructed to, and if fine-tuning only its language model, with the vision encoder frozen, installs the missing response (\cref{sec:finetune}).
\end{enumerate}

\section{Experiments}
\label{sec:experiments}
We evaluate ten Video LLMs on \name\ to answer the first research question, applying the checks of \cref{sec:metrics} in order after an overview of the main results.

\subsection{Experimental Setup}
\label{sec:setup}
\para{Models.} We evaluate two proprietary models, GPT-6.1 Terra and Gemini 3.8 Flash, and eight open-source models from five families (\cref{tab:main}). Within Qwen3-VL, three sizes (2B, 8B, 32B) and the reasoning variant of the 8B model isolate the effects of scale and reasoning.

\para{Protocol.} All models answer the main set under both prompts and, under the plain prompt, without video. Qwen3-VL-2B, Qwen3-VL-8B, and VideoLLaMA3-7B run the full perturbation set, and the proprietary models its conditions with a present, blank, or swapped video. Gemini receives the video file and GPT frames sampled at 1~fps (up to 32); open-source models use the frame sampling of their official examples. Open-source models decode greedily with 128 new tokens (256 for LLaVA-Video; 2,048 for the Thinking model, whose final answer alone is judged), and proprietary models use their API defaults with up to 1,024 output tokens (Supp.~\ref{supp:models}). The judge is DeepSeek-V4.1-Flash at temperature~0. Invalid responses stay below 2.5\% for every model and condition and are excluded.

\subsection{Main Results}
\label{sec:main}
\begin{figure}[t]
\centering
\includegraphics[width=\linewidth]{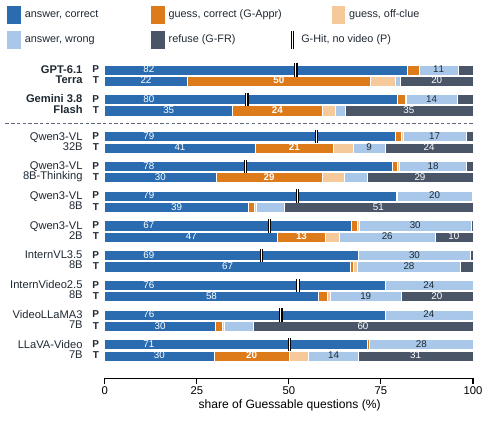}
\caption{\textbf{Responses to Guessable questions.} The two left segments add up to G-Hit and the dark orange one is G-Appr; the black tick marks G-Hit without the video. Under the plain prompt the clue-supported answer is stated as fact; the three-way prompt turns answers into refusals rather than guesses.}
\label{fig:dist}
\end{figure}

\cref{tab:main} reports all metrics, and \cref{fig:dist} breaks down the responses to \tG\ questions.

\para{By default, one response serves all three levels.} Under the plain prompt, every model answers most \tA\ questions correctly (A-Acc of at least 83.4\%) and treats the other two levels the same way: at least 92.8\% of \tG\ questions receive a definite answer, and no model guesses appropriately on more than 3.3\%. ES stays at or below 18.2, so the response type barely reveals whether an answer was seen, inferred, or invented.

\para{The three-way prompt installs a switch, not a scale.} Although the prompt offers all three responses, it mainly opens refusal, and refusal does not separate partial from absent evidence. U-Ref rises by 28 to 85 points, and G-FR rises with it: across the ten models, the Pearson correlation between G-FR and U-Ref is $r=\rUG$. The two models that refuse the most \tU\ questions, VideoLLaMA3-7B and Qwen3-VL-8B, also refuse 59.7\% and 51.3\% of \tG\ questions while guessing appropriately on 1.9\% and 1.6\%. ES rises only to 10.8--44.9. A perfect answer-or-refuse switch would score 61.6, and higher values require guessing: GPT-6.1 Terra, the only model that guesses appropriately on about half of the \tG\ questions (49.8\%), reaches the highest ES (44.9) and Avg (73.4\%), which shows that the middle response is attainable.

\subsection{Perception: Models See the Evidence}
\label{sec:see}
A model that does not see the evidence cannot be repaired by an instruction or a training target, so perception is checked first. All ten models pass.

\para{Answers and candidates follow the video.} Removing the video lowers A-Acc from 83.4--97.7\% to 23.8--42.2\%, and on the 178 \tA\ questions that no model answers correctly without the video, A-Acc with the video remains 75.3--97.2\% (Supp.~\ref{supp:main}), so accuracy is not carried by language priors. With the video, all five models in \cref{tab:interv} also accept a correct candidate more often than a wrong one by at least 64 points under the plain prompt, a gap that shrinks to at most 11 points without the video for the open-source models (Supp.~\ref{supp:interv}).

\para{The clue is seen, but its answer is asserted.} Averaged over the ten models, the clue-supported answer is reached on 76.8\% of \tG\ questions with the video and on 47.6\% without it (black ticks in \cref{fig:dist}), so the content draws on what the video shows. Yet only 1.1\% of \tG\ questions receive it as a guess. The gap between G-Hit and G-Appr, 66.9 to 82.2 points per model, is content that the video only suggests but the user receives as fact. The three-way prompt narrows this gap mainly by withholding the content rather than hedging it: for VideoLLaMA3-7B, G-Hit falls from 76.1\% to 32.0\% while G-Appr reaches only 1.9\%.

\begin{table}[t]
\centering
\footnotesize
\setlength{\tabcolsep}{1.8pt}
\resizebox{\linewidth}{!}{%
\begin{tabular}{@{}l cc cc cc cc cc@{}}
\toprule
 & \multicolumn{2}{c}{GPT-6.1} & \multicolumn{2}{c}{Gemini} & \multicolumn{2}{c}{Qwen-2B} & \multicolumn{2}{c}{Qwen-8B} & \multicolumn{2}{c}{VLLaMA3} \\
\cmidrule(lr){2-3}\cmidrule(lr){4-5}\cmidrule(lr){6-7}\cmidrule(lr){8-9}\cmidrule(l){10-11}
Perturbation & P & T & P & T & P & T & P & T & P & T \\
\midrule
\multicolumn{11}{@{}l}{\textit{Answerable questions (200), video present: accepted}}\\
Correct candidate\,$\uparrow$ & \textbf{98.5} & 97.0 & \underline{97.5} & 97.0 & 95.5 & \underline{98.5} & 94.5 & \textbf{99.0} & 97.0 & 95.5 \\
Wrong candidate\,$\downarrow$ & \underline{2.0} & \underline{1.5} & \textbf{1.0} & \textbf{0.5} & 31.0 & 39.5 & 5.5 & 30.5 & 29.6 & 38.5 \\
False premise\,$\downarrow$ & 46.7 & 46.2 & 40.7 & \textbf{13.1} & 37.5 & 71.0 & \textbf{19.1} & \underline{26.0} & \underline{31.0} & 29.1 \\
\midrule
\multicolumn{11}{@{}l}{\textit{Video perturbations (200 A): refused\,$\uparrow$}}\\
Blank video & \textbf{73.0} & \textbf{100.0} & \underline{43.0} & 98.0 & 0.0 & 55.1 & 2.5 & \underline{99.5} & 0.0 & 85.4 \\
Swapped video & \underline{58.9} & \textbf{95.9} & \textbf{66.8} & \underline{95.4} & 0.5 & 30.6 & 20.3 & \textbf{95.9} & 0.0 & 88.8 \\
\midrule
\multicolumn{11}{@{}l}{\textit{Question-text perturbations (100 U)}}\\
Original: refused\,$\uparrow$ & \textbf{40.0} & 72.0 & \underline{27.0} & 80.8 & 5.0 & 36.0 & 12.1 & \underline{81.0} & 3.0 & \textbf{85.0} \\
+ visibility claim: refused\,$\uparrow$ & \textbf{56.0} & \textbf{83.0} & \underline{40.8} & 77.0 & 13.1 & 38.0 & 13.3 & \underline{78.0} & 7.1 & \textbf{83.0} \\
+ wrong cand.: refused\,$\uparrow$ & \textbf{65.7} & \textbf{80.0} & \underline{39.0} & 74.7 & 2.0 & 30.0 & 0.0 & 66.0 & 0.0 & \underline{78.0} \\
+ wrong cand.: adopted\,$\downarrow$ & \textbf{6.1} & \textbf{5.0} & \underline{11.0} & \underline{5.1} & 73.0 & 58.0 & 19.0 & 30.0 & 70.7 & 22.0 \\
\bottomrule
\end{tabular}}
\caption{\textbf{Controlled perturbations (\%).} Arrows give the desired direction. Bold and underline rank the five models under the same prompt; conditions without video and the qualifier ladder are in Supp.~\ref{supp:interv}.}
\label{tab:interv}
\end{table}

\subsection{Judgment: Sufficiency Is Not Weighed}
\label{sec:judge}
Seeing the evidence does not imply weighing it. Low ES under the plain prompt (\cref{sec:main}) already fails the first part of the judgment check; the other two parts show what the response follows instead.

\para{Removing the evidence rarely changes the default response.} Without any video, eight of the ten models refuse at most 15.8\% of \tU\ questions under the plain prompt, although the prompt asks for an answer based only on the video. The proprietary models and Qwen3-VL-32B even refuse far less without the video than with it (GPT-6.1 Terra: 47.9\% to 5.2\%; Gemini 3.8 Flash: 39.4\% to 15.8\%; Qwen3-VL-32B: 23.4\% to 7.1\%): their refusals are triggered by inspecting a video that lacks the answer, not by the lack of evidence itself. With a blank or swapped video, the three open-source models refuse at most 2.5\% and 20.3\% of questions (\cref{tab:interv}); Qwen3-VL-8B, for instance, answers \emph{``guitar''} to the \tA\ question of \cref{fig:overview}a whether its video is present, removed, blank, or swapped.

\para{Synthetic absence is flagged; natural absence is not.} Without being asked, the proprietary models refuse 43.0--73.0\% of blank and 58.9--66.8\% of swapped videos, but only 27.0--40.0\% of the natural \tU\ questions of the perturbation set. The reasoning model shows the same split, refusing 89.2\% of blank-video questions but 26.5\% of natural \tU\ questions (Supp.~\ref{supp:interv}). A blank or mismatched clip is an anomaly in the input, whereas a natural video that shows the scene but not the answer requires judging sufficiency. Evaluations that blank or remove the input thus test the first ability and can overstate the second, which \name\ is built to measure.

\para{Wording that supplies content moves the response; wording that asserts evidence does not.} With the video fixed, a claim that the answer is clearly visible never lowers refusal of \tU\ questions by more than 4 points, and under the plain prompt it even raises refusal for all five models; up to four redundant qualifiers change A-Acc by at most two points (Supp.~\ref{supp:interv}). Content in the question acts differently: under the plain prompt, the open-source models adopt an embedded wrong candidate for 19.0--73.0\% of \tU\ questions, and all five models accept a false premise in 19.1--46.7\% of cases. The failure is therefore not deference to an assertive user but a readiness to state whatever content is at hand, from the video or from the question.

\begin{figure}[t]
\centering
\includegraphics[width=\linewidth]{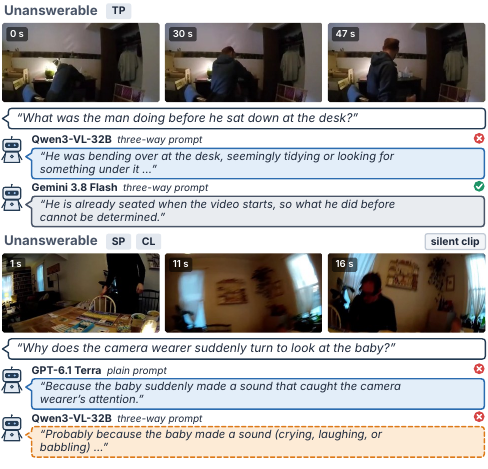}
\caption{\textbf{Answers without evidence.} Top: the man is already at the desk when the clip starts (TP), yet Qwen3-VL-32B describes what he did before; Gemini 3.8 Flash notices the gap. Bottom: the reason for the head turn is out of frame (SP) and the clip is silent, yet both responses explain it by a sound. Responses translated and abridged.}
\label{fig:examples}
\end{figure}

\para{The check follows visible traces, not sufficiency.} Under the three-way prompt, refusal depends on why the answer is missing (\cref{tab:main}, right). A temporal gap leaves no trace in the frames, unlike an occluder or a blurred target, and seven of the eight open-source models refuse TP questions least (InternVL3.5-8B: 56.0\% for PI versus 13.6\% for TP), so the gap that only video poses is where they most often answer what they cannot see. Causal questions are refused less often than other \tU\ questions by every model, by 3 to 61 points (Qwen3-VL-8B-Thinking: 27.0\% versus 80.0\%), and attract the rare guesses about as often without a clue as with one (Supp.~\ref{supp:main}). \cref{fig:examples} shows both patterns. In sum, the evidence determines \emph{what} a model says, while the instruction and the form of the question determine \emph{how} it responds.

\subsection{Capacity: Instruction, Scale, and Reasoning}
\label{sec:scale}
A judgment failure can reflect missing capacity, which larger or reasoning models should remedy, or missing supervision, which an instruction can expose but not install.

\para{The check exists, but only on request.} Under the three-way prompt, the five models refuse 55.1--100\% of blank-video and 30.6--95.9\% of swapped-video questions, and the open-source models refuse every \tU\ question asked without video (Supp.~\ref{supp:interv}). Detecting missing evidence is within their reach once refusal is licensed; what they lack is a response for evidence that is incomplete but informative.

\para{Scale and reasoning raise refusal without grading the evidence.} From 2B to 32B, Qwen3-VL raises G-Hit from 68.5\% to 80.6\% and plain-prompt U-Ref from 6.1\% to 23.4\%, and under the three-way prompt the 32B and reasoning models guess more often than the 8B model (G-Appr 21.3\% and 28.8\% versus 1.6\%). Yet their ES (33.4 and 34.5) stays close to that of the 8B model (31.1). Larger and reasoning models see more and refuse more, but neither scale nor reasoning supplies the graded response, which points to missing supervision; \cref{sec:finetune} tests this directly.

\subsection{Diagnosis across Models}
\label{sec:diag}
Applied in order, the three checks give all ten models the same diagnosis. Every model passes the perception check: removing the video costs at least 49.6 points of A-Acc and 21.8 points of G-Hit. Every model fails the judgment check: its plain-prompt ES is at most 18.2, and without the video it still refuses fewer than half of the \tU\ questions. Models differ only in how far the default check reaches: the proprietary and reasoning models refuse 43.0--89.2\% of blank clips and the other three tested models at most 2.5\%, yet none guesses appropriately on more than 3.3\% of the \tG\ questions. The diagnosis also holds within each video source, where plain-prompt G-Appr stays at most 3.3\% (Supp.~\ref{supp:main}). The third check places the failure in supervision, so the remedy is a training target rather than a stronger visual encoder, which \cref{sec:finetune} tests.

\section{Fine-Tuning Validation}
\label{sec:finetune}
This section applies the third check of \cref{sec:metrics} and answers the second research question: does supervising the missing guess level repair the failure where conventional refusal tuning does not? A positive answer both confirms the diagnosis and yields a model that responds to partial evidence.

\subsection{Setup and Hypotheses}
\label{sec:ftdesign}
\para{Data.} We annotate 608 new questions on 228 Charades videos with the guidelines of \cref{sec:build}: 546 for training (205 \tA, 147 \tG, 194 \tU) and 62 for validation. The videos come from actors who do not appear in the benchmark, and no Ego4D video is used, so the Ego4D half of \name\ tests transfer across sources.

\para{Targets.} The input is the video, the three-way prompt, and the question. \emph{Refusal-aware} targets, the usual recipe of refusal tuning~\cite{zhang2024rtuning}, give the answer for \tA\ questions and \emph{``Cannot be determined from the video.''} for both other levels. \emph{Guess-aware} targets differ only on \tG\ questions, which receive \emph{``Probably X, because C.''} with the reference guess X and its clue C, so that the guess can be checked. The base model sees the same prompt, which already offers a hedged guess.

\para{Training.} We train low-rank adapters~\cite{hu2022lora} on the language model of Qwen3-VL-8B with the vision encoder frozen (Supp.~\ref{supp:ft}), so any gain in hedged guessing comes from how the language model responds to clues it already receives. The two runs differ only in the 147 \tG\ targets.

\para{Hypotheses.} \textbf{H1}: guess-aware targets raise G-Appr without raising false refusals. \textbf{H2}: refusal-aware targets, which merge partial with absent evidence, raise refusal on \tG\ \emph{and} \tA\ questions.

\begin{table}[t]
\centering
\footnotesize
\setlength{\tabcolsep}{3.0pt}
\resizebox{\linewidth}{!}{%
\begin{tabular}{@{}l r@{\,}l r@{\,}l r@{\,}l@{}}
\toprule
 & \multicolumn{2}{c}{Refusal-aware} & \multicolumn{2}{c}{Guess-aware} & \multicolumn{2}{c}{Guess-aware$^{\ddagger}$} \\
\midrule
\multicolumn{7}{@{}l}{\textit{Paired change vs.\ base (three-way prompt), 95\% CI}}\\
A-Acc & $-$10.4 & {\scriptsize[$-$13.5, $-$7.4]} & \underline{$-$5.9} & {\scriptsize[$-$8.7, $-$3.1]} & \textbf{$-$2.3} & {\scriptsize[$-$5.1, 0.5]} \\
A-FR\,$\downarrow$ & \textcolor{ftbad}{$+$13.0} & {\scriptsize[10.4, 15.8]} & \underline{$-$0.6} & {\scriptsize[$-$2.0, 0.8]} & \textbf{\textcolor{ftgood}{$-$1.7}} & {\scriptsize[$-$2.8, $-$0.6]} \\
G-Appr & \textcolor{ftbad}{$-$1.4} & {\scriptsize[$-$2.6, $-$0.5]} & \underline{\textcolor{ftgood}{$+$39.0}} & {\scriptsize[34.5, 43.6]} & \textbf{\textcolor{ftgood}{$+$48.8}} & {\scriptsize[44.1, 53.3]} \\
G-FR\,$\downarrow$ & \textcolor{ftbad}{$+$41.3} & {\scriptsize[36.8, 45.7]} & \underline{\textcolor{ftgood}{$-$19.6}} & {\scriptsize[$-$24.5, $-$14.6]} & \textbf{\textcolor{ftgood}{$-$32.9}} & {\scriptsize[$-$38.6, $-$27.4]} \\
U-Ref & \textbf{$+$12.5} & {\scriptsize[10.0, 15.2]} & \underline{$+$0.6} & {\scriptsize[$-$2.1, 3.3]} & $-$6.5 & {\scriptsize[$-$9.9, $-$3.2]} \\
\midrule
\textit{Transfer and retention} & \multicolumn{2}{c}{\textit{Base}} & \multicolumn{2}{c}{\textit{Guess-aware}} & \multicolumn{2}{c}{\textit{Guess-aware}$^{\ddagger}$} \\
\cmidrule(lr){2-3}\cmidrule(lr){4-5}\cmidrule(l){6-7}
G-Appr, Ego4D & \multicolumn{2}{c}{1.6} & \multicolumn{2}{c}{41.5} & \multicolumn{2}{c}{51.1} \\
Blank refusal & \multicolumn{2}{c}{99.5} & \multicolumn{2}{c}{95.0} & \multicolumn{2}{c}{63.5} \\
Video-MME & \multicolumn{2}{c}{68.8} & \multicolumn{2}{c}{67.2} & \multicolumn{2}{c}{69.4} \\
\bottomrule
\end{tabular}}
\caption{\textbf{Fine-tuning Qwen3-VL-8B (\%).} Top: change from the base model under the three-way prompt, paired by question, with 95\% bootstrap intervals over videos; green and red mark significant improvements and deteriorations on the metrics of H1 and H2. Bold: best; underlined: second best. Bottom: G-Appr on Ego4D, which contributes no training video; refusal of blank videos; accuracy on a fixed 500-question subset of Video-MME~\cite{fu2025videomme}. The base model is re-evaluated in the same pipeline. $^{\ddagger}$Trained with both prompts.}
\label{tab:ft}
\end{table}

\begin{figure}[t]
\centering
\includegraphics[width=\linewidth]{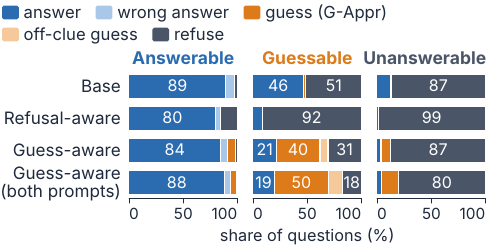}
\caption{\textbf{Response types after fine-tuning} (three-way prompt). Refusal-aware targets turn Guessable questions into refusals and spill onto Answerable ones; guess-aware targets open the guess band on Guessable questions while the other two levels keep their response.}
\label{fig:ft}
\end{figure}

\subsection{Results}
\label{sec:ftresults}
\para{H1 holds: guess-aware supervision opens the guess level.} With 147 guess targets, G-Appr rises from 1.6\% to 40.4\% (\cref{tab:main}), a paired gain of 39.0 points [34.5, 43.6] over the base model re-evaluated in the same pipeline (\cref{tab:ft}), while A-FR ($-0.6$ [$-2.0$, 0.8]) and U-Ref (87.3\%) do not move. The new guesses replace refusals (81 of 173) and assertions (85) alike, and they land where the clues are: 47.8\% of \tG\ questions receive a guess, against 7.3\% of \tA\ and 9.2\% of \tU\ ones (\cref{fig:ft}); the A-Acc change comes from these hedged answers, not from refusals. Trained only on Charades, the model raises G-Appr on Ego4D from 1.6\% to 41.5\%, so the learned response follows partial evidence rather than familiar videos. That a rank-16 adapter on a frozen vision encoder suffices indicates that the base model already separates partial from absent evidence and lacks only the supervision to express it.

\para{H2 holds: refusal-aware supervision spreads refusal.} Refusal-aware tuning raises U-Ref to 99.2\%, but it also refuses 91.6\% of \tG\ questions and raises false refusals on \tA\ ones from 2.7\% to 15.7\% (paired increase 13.0 points [10.4, 15.8]). Folding partial evidence into the refusal class teaches that incompleteness of any degree warrants refusal, the same switch that the three-way prompt induces (\cref{sec:main}). Despite near-perfect refusal, its ES reaches only 36.1, against 43.9 for guess-aware tuning.

\subsection{Transfer Across Prompts}
\label{sec:ftlimits}
\para{Training with both prompts makes guessing prompt-independent.} Training each example with both prompts raises G-Appr under the plain prompt to 37.9\%, from 11.7\% with training on the three-way prompt alone, and under the three-way prompt to 50.2\%, the highest of all models in \cref{tab:main}. The same model keeps A-FR at 1.1\%, reaches an ES of 48.5 and 51.1\% G-Appr on Ego4D, and stays within two points of the base model on Video-MME. Guessing also stays on partial evidence: the guess-aware model still refuses 95.0\% of blank-video questions (\cref{tab:ft}).

\section{Conclusion}
\label{sec:conclusion}
We present \name, a diagnostic benchmark that asks whether Video LLMs match their response to the evidence a video provides. By separating partial from absent evidence and content from form, and by changing one input at a time, \name\ turns an accuracy or refusal rate into a diagnosis of where a failure lies. Across ten Video LLMs, the diagnosis places the failure in judgment rather than perception: models reach the clue-supported answer on most Guessable questions yet state it as fact, flag blank or mismatched clips more readily than natural gaps, and once instructed to abstain, refuse partial and absent evidence alike. The diagnosis is also actionable. Supervising the missing guess level, which changes only 147 training targets, raises appropriate guessing from 1.6\% to 40.4\% without extra false refusals and transfers to an unseen video source, whereas conventional refusal tuning spreads refusal to questions the video answers. \name\ thus points evaluation and training to the same target: not whether a model refuses, but whether its response matches the evidence, which calls for natural gaps rather than removed inputs in testing, and for a guess level rather than refusal alone in training. We will release the benchmark and code and extend the diagnosis to long and streaming videos.

\section*{LLM Usage Statement}
We used the LLM assistants Claude (Anthropic), Gemini (Google), and ChatGPT (OpenAI) to draft and edit the text of all sections, to write the code that draws the figures and builds the tables from our result files, and to cross-check the numbers in the text against those files. The benchmark design, the experiments, the analyses, and the conclusions are the authors' own, and the authors have checked all content and take full responsibility for it.

{
    \small
    \bibliographystyle{ieeenat_fullname}
    \bibliography{main}
}
\clearpage
\maketitlesupplementary
\appendix
\renewcommand{\thetable}{S\arabic{table}}
\renewcommand{\thefigure}{S\arabic{figure}}
\setcounter{table}{0}
\setcounter{figure}{0}
\renewcommand{\theHfigure}{S\arabic{figure}}
\renewcommand{\theHtable}{S\arabic{table}}
\noindent This supplement follows the order of the main paper. \cref{tab:nav} lists every place where the paper points to the supplement and where that material is found. Beyond these pointers, \cref{supp:second,supp:star,supp:fewshot,supp:calib} add four experiments on the fine-tuned models, and \cref{supp:qual} collects qualitative examples. In every table, the best value among the rows that share a condition is in \textbf{bold} and the second best is \underline{underlined}; columns without a preferred direction are left unmarked. Questions, references, and model responses in the figures and tables are presented in English for readability and abridged where marked.

\begin{table}[h]
\centering
\scriptsize
\setlength{\tabcolsep}{3pt}
\renewcommand{\arraystretch}{1.12}
\begin{tabular}{@{}p{0.115\linewidth}p{0.40\linewidth}p{0.42\linewidth}@{}}
\toprule
Paper & Content & Here \\
\midrule
Sec.~3.1 & Evidence levels and gap tags, with examples & \cref{supp:levels}; \cref{fig:gallery} \\
Sec.~3.2 & Annotation rules, quality control & \cref{supp:guide}; \cref{fig:flow} \\
Sec.~3.2 & Statistics; causal questions by source & \cref{tab:stats}; \cref{fig:stats} \\
Sec.~3.2 & Perturbation set, one example in full & \cref{supp:construct}; \cref{fig:family} \\
Sec.~3.3 & Judge agreement ($\kappa=0.96$, $0.84$) & \cref{supp:prompts}; \cref{fig:judge} \\
Sec.~3.3 & ES of at most 61.6 without guessing & \cref{supp:esbound} \\
Sec.~4.1 & Prompts, inference settings & \cref{supp:prompts,supp:models}; \cref{tab:models} \\
Sec.~4.2 & All response types at all levels & \cref{fig:distall}; \cref{tab:dist} \\
Sec.~4.3 & 178 vision-necessary questions & \cref{tab:vn} \\
Sec.~4.3 & Candidate check, with and without video & \cref{fig:cand}; \cref{tab:cand,tab:induce} \\
Sec.~4.4 & Gap tags; causal questions & \cref{fig:gapmap}; \cref{tab:gap,tab:causal} \\
Sec.~4.4 & Visibility claim; redundant qualifiers; reasoning model & \cref{tab:induce,tab:ladder}; \cref{supp:pres} \\
Sec.~4.6 & Diagnosis within each video source & \cref{fig:source}; \cref{tab:source} \\
Sec.~5.1 & Training data, targets, settings & \cref{fig:targets}; \cref{tab:fthp} \\
Sec.~5.2 & Results under both prompts and by source & \cref{tab:ftfull,tab:ftsrc} \\
\midrule
\multicolumn{2}{@{}l}{Second base model} & \cref{supp:second}; \cref{fig:ft2b} \\
\multicolumn{2}{@{}l}{Questions from an external benchmark} & \cref{supp:star}; \cref{fig:star} \\
\multicolumn{2}{@{}l}{Few-shot demonstrations as a baseline} & \cref{supp:fewshot}; \cref{tab:fewshot} \\
\multicolumn{2}{@{}l}{Reliability of each response type} & \cref{supp:calib}; \cref{fig:calib} \\
\multicolumn{2}{@{}l}{Qualitative examples} & \cref{supp:qual}; \cref{fig:wall,fig:wall2,fig:beforeafter} \\
\bottomrule
\end{tabular}
\caption{\textbf{Where to find what.} Upper part: pointers from the main paper. Lower part: additional experiments and examples.}
\label{tab:nav}
\end{table}

\section{Benchmark Details}
\label{supp:data}

\subsection{Evidence Levels and Gap Tags in Practice}\label{supp:levels}

\begin{figure*}[t]
\centering
\begin{minipage}[t]{0.492\textwidth}\vspace{0pt}
\begin{tikzpicture}[inner sep=0]\node(a){\includegraphics[width=0.326\linewidth]{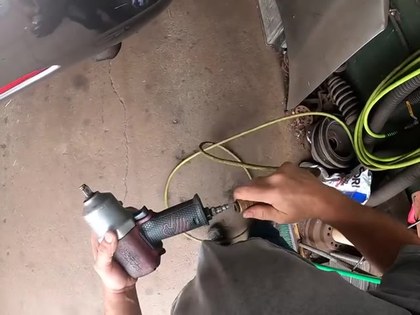}};\node[anchor=south west,fill=black,fill opacity=.55,text opacity=1,text=white,font=\tiny\sffamily,inner sep=1.2pt] at (a.south west){3 s};\end{tikzpicture}\hfill\begin{tikzpicture}[inner sep=0]\node(a){\includegraphics[width=0.326\linewidth]{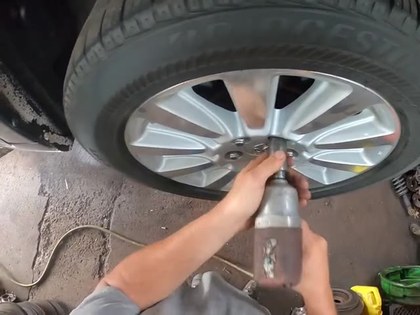}};\node[anchor=south west,fill=black,fill opacity=.55,text opacity=1,text=white,font=\tiny\sffamily,inner sep=1.2pt] at (a.south west){25 s};\end{tikzpicture}\hfill\begin{tikzpicture}[inner sep=0]\node(a){\includegraphics[width=0.326\linewidth]{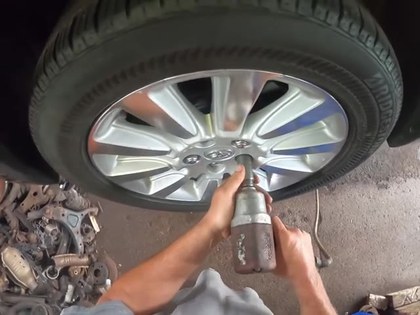}};\node[anchor=south west,fill=black,fill opacity=.55,text opacity=1,text=white,font=\tiny\sffamily,inner sep=1.2pt] at (a.south west){47 s};\end{tikzpicture}\\[2pt]
{\footnotesize \lvA\\[1pt]\textbf{Q:} \emph{``What type of tool does the camera wearer use to tighten the wheel nuts?''}\\[1pt]\textbf{Reference answer:} a pneumatic impact wrench. The tool is in view at the wheel.}
\end{minipage}\hfill
\begin{minipage}[t]{0.492\textwidth}\vspace{0pt}
\begin{tikzpicture}[inner sep=0]\node(a){\includegraphics[width=0.326\linewidth]{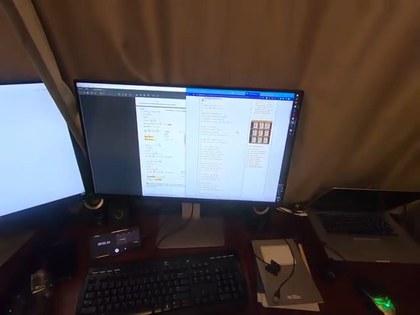}};\node[anchor=south west,fill=black,fill opacity=.55,text opacity=1,text=white,font=\tiny\sffamily,inner sep=1.2pt] at (a.south west){4 s};\end{tikzpicture}\hfill\begin{tikzpicture}[inner sep=0]\node(a){\includegraphics[width=0.326\linewidth]{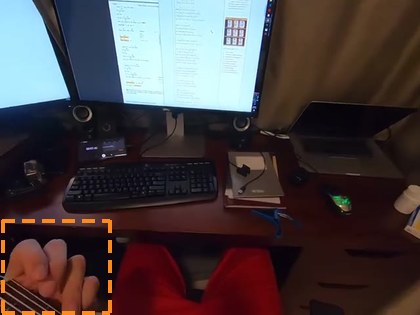}};\node[anchor=south west,fill=black,fill opacity=.55,text opacity=1,text=white,font=\tiny\sffamily,inner sep=1.2pt] at (a.south west){9 s};\end{tikzpicture}\hfill\begin{tikzpicture}[inner sep=0]\node(a){\includegraphics[width=0.326\linewidth]{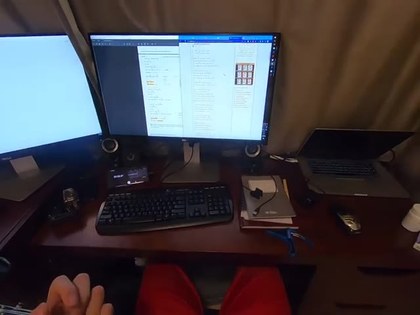}};\node[anchor=south west,fill=black,fill opacity=.55,text opacity=1,text=white,font=\tiny\sffamily,inner sep=1.2pt] at (a.south west){10 s};\end{tikzpicture}\\[2pt]
{\footnotesize \lvG\ \stag{SP}\\[1pt]\textbf{Q:} \emph{``What type of instrument is the object in the wearer's hand, of which only a small part is in view?''}\\[1pt]\textbf{Reference guess:} a guitar; a bass; a ukulele. \textbf{Clue:} a dark neck with parallel strings at the lower left (dashed box), with the wearer's fingers pressing the strings. The body stays out of frame.}
\end{minipage}\\[7pt]
\begin{minipage}[t]{0.492\textwidth}\vspace{0pt}
\begin{tikzpicture}[inner sep=0]\node(a){\includegraphics[width=0.326\linewidth]{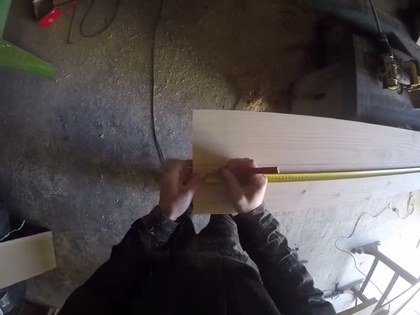}};\node[anchor=south west,fill=black,fill opacity=.55,text opacity=1,text=white,font=\tiny\sffamily,inner sep=1.2pt] at (a.south west){2 s};\end{tikzpicture}\hfill\begin{tikzpicture}[inner sep=0]\node(a){\includegraphics[width=0.326\linewidth]{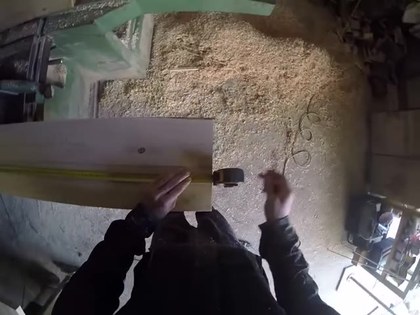}};\node[anchor=south west,fill=black,fill opacity=.55,text opacity=1,text=white,font=\tiny\sffamily,inner sep=1.2pt] at (a.south west){9 s};\end{tikzpicture}\hfill\begin{tikzpicture}[inner sep=0]\node(a){\includegraphics[width=0.326\linewidth]{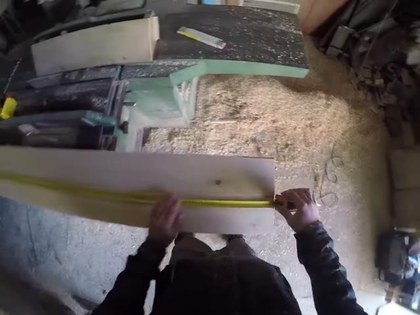}};\node[anchor=south west,fill=black,fill opacity=.55,text opacity=1,text=white,font=\tiny\sffamily,inner sep=1.2pt] at (a.south west){15 s};\end{tikzpicture}\\[2pt]
{\footnotesize \lvG\ \stag{PI}\\[1pt]\textbf{Q:} \emph{``What type of tool is the thin red object in the wearer's right hand?''}\\[1pt]\textbf{Reference guess:} a pencil. \textbf{Clue:} the wearer holds it along a tape measure and marks the board with it. The object is too small to identify by its appearance.}
\end{minipage}\hfill
\begin{minipage}[t]{0.492\textwidth}\vspace{0pt}
\begin{tikzpicture}[inner sep=0]\node(a){\includegraphics[width=0.326\linewidth]{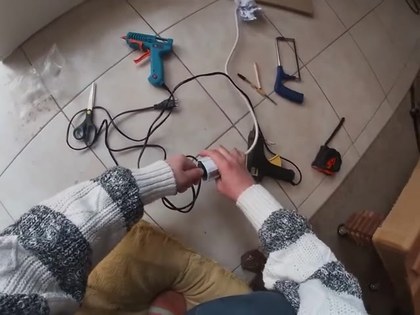}};\node[anchor=south west,fill=black,fill opacity=.55,text opacity=1,text=white,font=\tiny\sffamily,inner sep=1.2pt] at (a.south west){11 s};\end{tikzpicture}\hfill\begin{tikzpicture}[inner sep=0]\node(a){\includegraphics[width=0.326\linewidth]{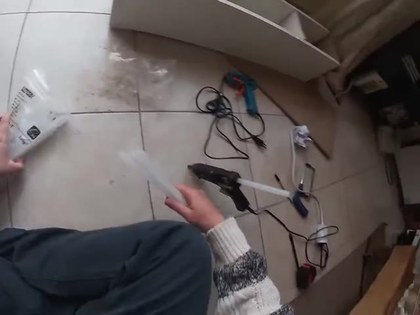}};\node[anchor=south west,fill=black,fill opacity=.55,text opacity=1,text=white,font=\tiny\sffamily,inner sep=1.2pt] at (a.south west){36 s};\end{tikzpicture}\hfill\begin{tikzpicture}[inner sep=0]\node(a){\includegraphics[width=0.326\linewidth]{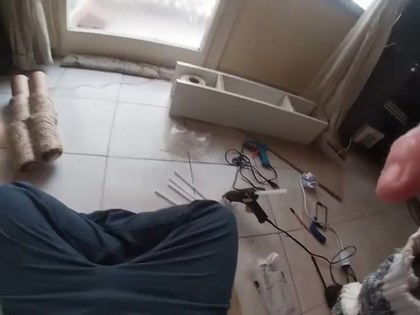}};\node[anchor=south west,fill=black,fill opacity=.55,text opacity=1,text=white,font=\tiny\sffamily,inner sep=1.2pt] at (a.south west){41 s};\end{tikzpicture}\\[2pt]
{\footnotesize \lvG\ \stag{TP}\\[1pt]\textbf{Q:} \emph{``After taking the white glue sticks out of the plastic bag, what will the wearer do next?''}\\[1pt]\textbf{Reference guess:} load a stick into the hot glue gun; glue the twine onto the paper tube. \textbf{Clue:} a plugged-in glue gun on the floor and a paper tube wrapped with twine. The clip ends before the next step.}
\end{minipage}\\[7pt]
\begin{minipage}[t]{0.492\textwidth}\vspace{0pt}
\begin{tikzpicture}[inner sep=0]\node(a){\includegraphics[width=0.326\linewidth]{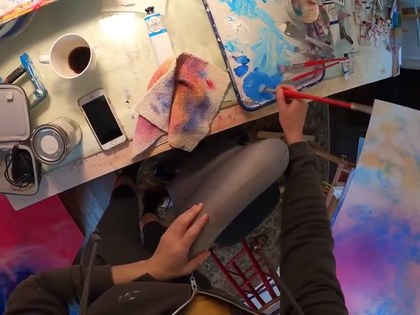}};\node[anchor=south west,fill=black,fill opacity=.55,text opacity=1,text=white,font=\tiny\sffamily,inner sep=1.2pt] at (a.south west){2 s};\end{tikzpicture}\hfill\begin{tikzpicture}[inner sep=0]\node(a){\includegraphics[width=0.326\linewidth]{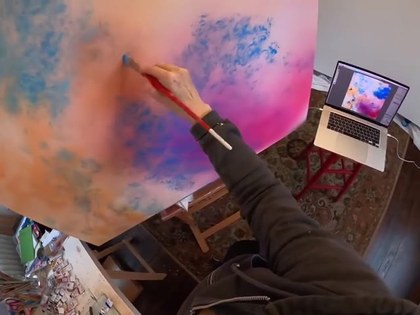}};\node[anchor=south west,fill=black,fill opacity=.55,text opacity=1,text=white,font=\tiny\sffamily,inner sep=1.2pt] at (a.south west){16 s};\end{tikzpicture}\hfill\begin{tikzpicture}[inner sep=0]\node(a){\includegraphics[width=0.326\linewidth]{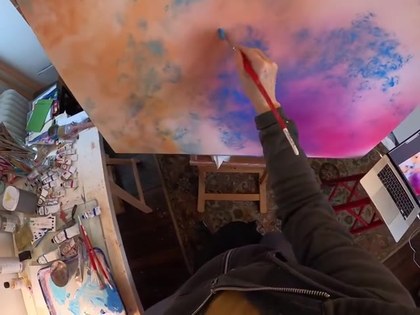}};\node[anchor=south west,fill=black,fill opacity=.55,text opacity=1,text=white,font=\tiny\sffamily,inner sep=1.2pt] at (a.south west){30 s};\end{tikzpicture}\\[2pt]
{\footnotesize \lvU\ \stag{SP}\\[1pt]\textbf{Q:} \emph{``What color is the camera wearer's hair?''}\\[1pt]\textbf{Missing evidence:} the head-mounted camera never shows the wearer's head.}
\end{minipage}\hfill
\begin{minipage}[t]{0.492\textwidth}\vspace{0pt}
\begin{tikzpicture}[inner sep=0]\node(a){\includegraphics[width=0.326\linewidth]{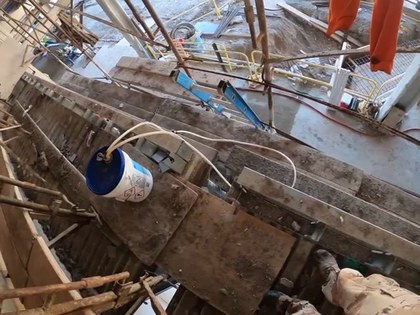}};\node[anchor=south west,fill=black,fill opacity=.55,text opacity=1,text=white,font=\tiny\sffamily,inner sep=1.2pt] at (a.south west){1 s};\end{tikzpicture}\hfill\begin{tikzpicture}[inner sep=0]\node(a){\includegraphics[width=0.326\linewidth]{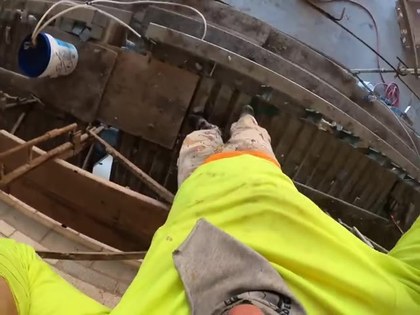}};\node[anchor=south west,fill=black,fill opacity=.55,text opacity=1,text=white,font=\tiny\sffamily,inner sep=1.2pt] at (a.south west){3 s};\end{tikzpicture}\hfill\begin{tikzpicture}[inner sep=0]\node(a){\includegraphics[width=0.326\linewidth]{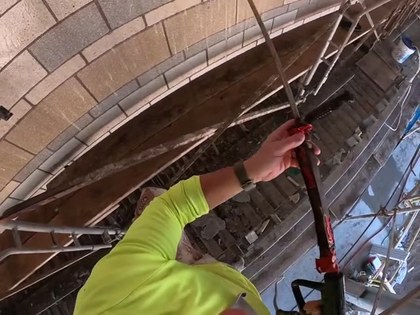}};\node[anchor=south west,fill=black,fill opacity=.55,text opacity=1,text=white,font=\tiny\sffamily,inner sep=1.2pt] at (a.south west){31 s};\end{tikzpicture}\\[2pt]
{\footnotesize \lvU\ \stag{PI}\\[1pt]\textbf{Q:} \emph{``What product name is printed on the label of the white bucket with the blue lid on the scaffold?''}\\[1pt]\textbf{Missing evidence:} the bucket and its label are in view, and the print is far too small to read.}
\end{minipage}\\[7pt]
\begin{minipage}[t]{0.492\textwidth}\vspace{0pt}
\begin{tikzpicture}[inner sep=0]\node(a){\includegraphics[width=0.326\linewidth]{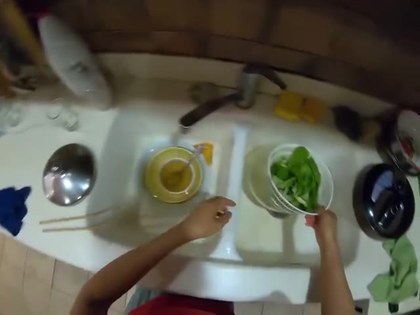}};\node[anchor=south west,fill=black,fill opacity=.55,text opacity=1,text=white,font=\tiny\sffamily,inner sep=1.2pt] at (a.south west){2 s};\end{tikzpicture}\hfill\begin{tikzpicture}[inner sep=0]\node(a){\includegraphics[width=0.326\linewidth]{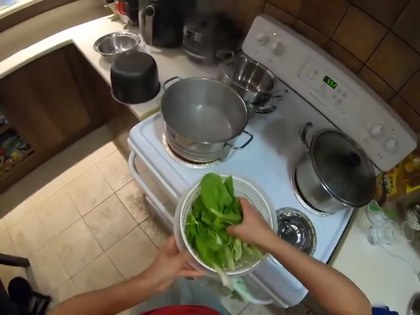}};\node[anchor=south west,fill=black,fill opacity=.55,text opacity=1,text=white,font=\tiny\sffamily,inner sep=1.2pt] at (a.south west){8 s};\end{tikzpicture}\hfill\begin{tikzpicture}[inner sep=0]\node(a){\includegraphics[width=0.326\linewidth]{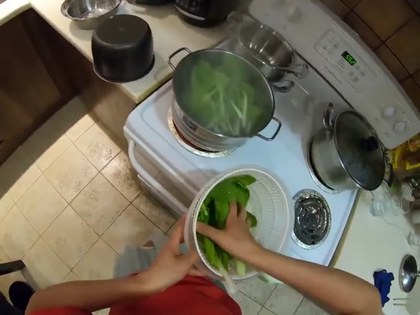}};\node[anchor=south west,fill=black,fill opacity=.55,text opacity=1,text=white,font=\tiny\sffamily,inner sep=1.2pt] at (a.south west){14 s};\end{tikzpicture}\\[2pt]
{\footnotesize \lvU\ \stag{TP}\\[1pt]\textbf{Q:} \emph{``For how long will the wearer boil these greens?''}\\[1pt]\textbf{Missing evidence:} the cooking time lies after the clip, and nothing in view indicates it.}
\end{minipage}\hfill
\begin{minipage}[t]{0.492\textwidth}\vspace{0pt}
\begin{tikzpicture}[inner sep=0]\node(a){\includegraphics[width=0.326\linewidth]{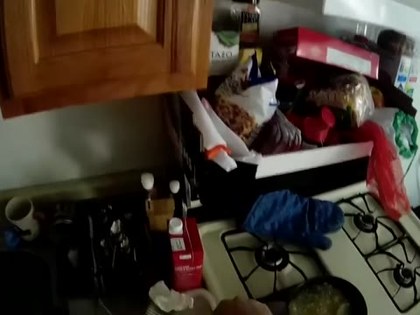}};\node[anchor=south west,fill=black,fill opacity=.55,text opacity=1,text=white,font=\tiny\sffamily,inner sep=1.2pt] at (a.south west){5 s};\end{tikzpicture}\hfill\begin{tikzpicture}[inner sep=0]\node(a){\includegraphics[width=0.326\linewidth]{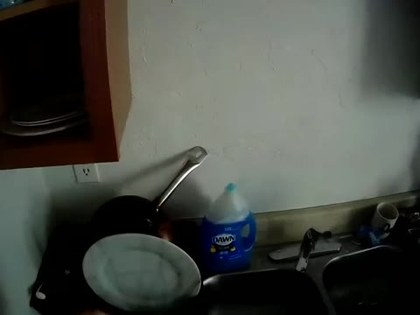}};\node[anchor=south west,fill=black,fill opacity=.55,text opacity=1,text=white,font=\tiny\sffamily,inner sep=1.2pt] at (a.south west){12 s};\end{tikzpicture}\hfill\begin{tikzpicture}[inner sep=0]\node(a){\includegraphics[width=0.326\linewidth]{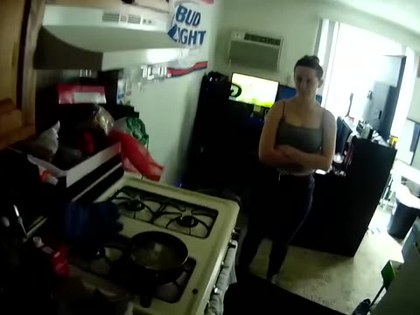}};\node[anchor=south west,fill=black,fill opacity=.55,text opacity=1,text=white,font=\tiny\sffamily,inner sep=1.2pt] at (a.south west){16 s};\end{tikzpicture}\\[2pt]
{\footnotesize \lvU\ \stag{SP}\ \stag{CL}\\[1pt]\textbf{Q:} \emph{``Why is the woman standing by the stove with her arms crossed, looking at the camera wearer?''}\\[1pt]\textbf{Missing evidence:} the reason depends on what the wearer is doing off camera, and the clip is silent.}
\end{minipage}
\caption{\textbf{One question per evidence level and gap tag.} Frames are taken from the clip at the times shown. Guessable questions record every candidate that the clue supports, together with the clue; Unanswerable questions record which evidence is missing. SP: spatial gap; PI: perceptual gap; TP: temporal gap; CL: causal question.}
\label{fig:gallery}
\end{figure*}

\cref{fig:gallery} shows one real question for each combination of evidence level and gap tag. The three Guessable questions share one property: the answer itself is never in view, yet a visible object, state, or action that any viewer can point to narrows it to a few candidates. In the first, only the neck of an instrument enters the frame, and the strings under the wearer's fingers limit the answer to a guitar, a bass, or a ukulele. In the third, the clip ends before the next step, and the glue gun that the wearer has just plugged in indicates what the glue sticks are for. The Unanswerable questions in the bottom rows look similar on the surface, since they also ask about a hidden, tiny, or future target, yet the frames offer nothing that favors one answer over another. A binary label would put both groups in the same class. Separating them is what allows ClueProbe to score a hedged, clue-based response as correct on the first group and a refusal as correct on the second.

\subsection{Annotation Guidelines}\label{supp:guide}

\para{Question rules.} A question must (i) refer to a single target that its wording identifies uniquely in the video, (ii) not reveal or strongly suggest its answer, (iii) not presuppose an event the video does not show, (iv) ask at a granularity at which the level is well defined, and (v) be decidable from the silent video alone, without audio or outside knowledge of the recording. Rule (iv) matters because the level depends on what is asked: asking for the \emph{type} of an object can be Guessable where asking for its brand is Unanswerable.

\begin{figure}[t]
\centering
\begin{tikzpicture}[node distance=7pt and 8pt]
\tikzset{q/.style={flowbox,font=\footnotesize,text width=4.25cm},o/.style={rounded corners=2pt,font=\footnotesize\sffamily\bfseries,text=white,inner sep=3pt,text width=2.05cm,align=center},rec/.style={font=\scriptsize,align=left,text width=2.2cm,inner sep=1pt,text=cInk}}
\node[q](q1){\textbf{Step 1.} Do the frames confirm one answer at the granularity asked?};
\node[o,fill=cA,right=13pt of q1](a){Answerable};
\node[rec,below=1pt of a]{reference answer};
\node[q,below=24pt of q1](q2){\textbf{Step 2.} Does a visible clue, which a second viewer can also find, narrow the answer to a few candidates?};
\node[o,fill=cG,right=13pt of q2](g){Guessable};
\node[rec,below=1pt of g]{all supported candidates, the clue, gap tag};
\node[q,below=30pt of q2](q3){\textbf{Step 3.} Write down which evidence is missing, and why it is not visible.};
\node[o,fill=cU,right=13pt of q3](u){Unanswerable};
\node[rec,below=1pt of u]{missing evidence, an unsupported answer, gap tag};
\draw[arr](q1)--node[above,font=\scriptsize]{yes}(a);\draw[arr](q2)--node[above,font=\scriptsize]{yes}(g);\draw[arr](q3)--(u);
\draw[arr](q1)--node[right,font=\scriptsize]{no}(q2);\draw[arr](q2)--node[right,font=\scriptsize]{no}(q3);
\node[softbox,below=30pt of q3.south west,anchor=north west,font=\footnotesize,text width=7.5cm](t){\textbf{Tags.} Gap, for Guessable and Unanswerable: \stag{SP} out of frame or occluded, \stag{PI} in view but too small or blurred, \stag{TP} before or after the clip. Question logic, at every level: \stag{CL} asks why something happened.};
\end{tikzpicture}
\caption{\textbf{How a question receives its level.} The steps are applied in order, using the video alone; the small text under each level lists what the annotation records. The clue of a Guessable question must be a concrete visible fact, so that a guess can be checked against it.}
\label{fig:flow}
\end{figure}
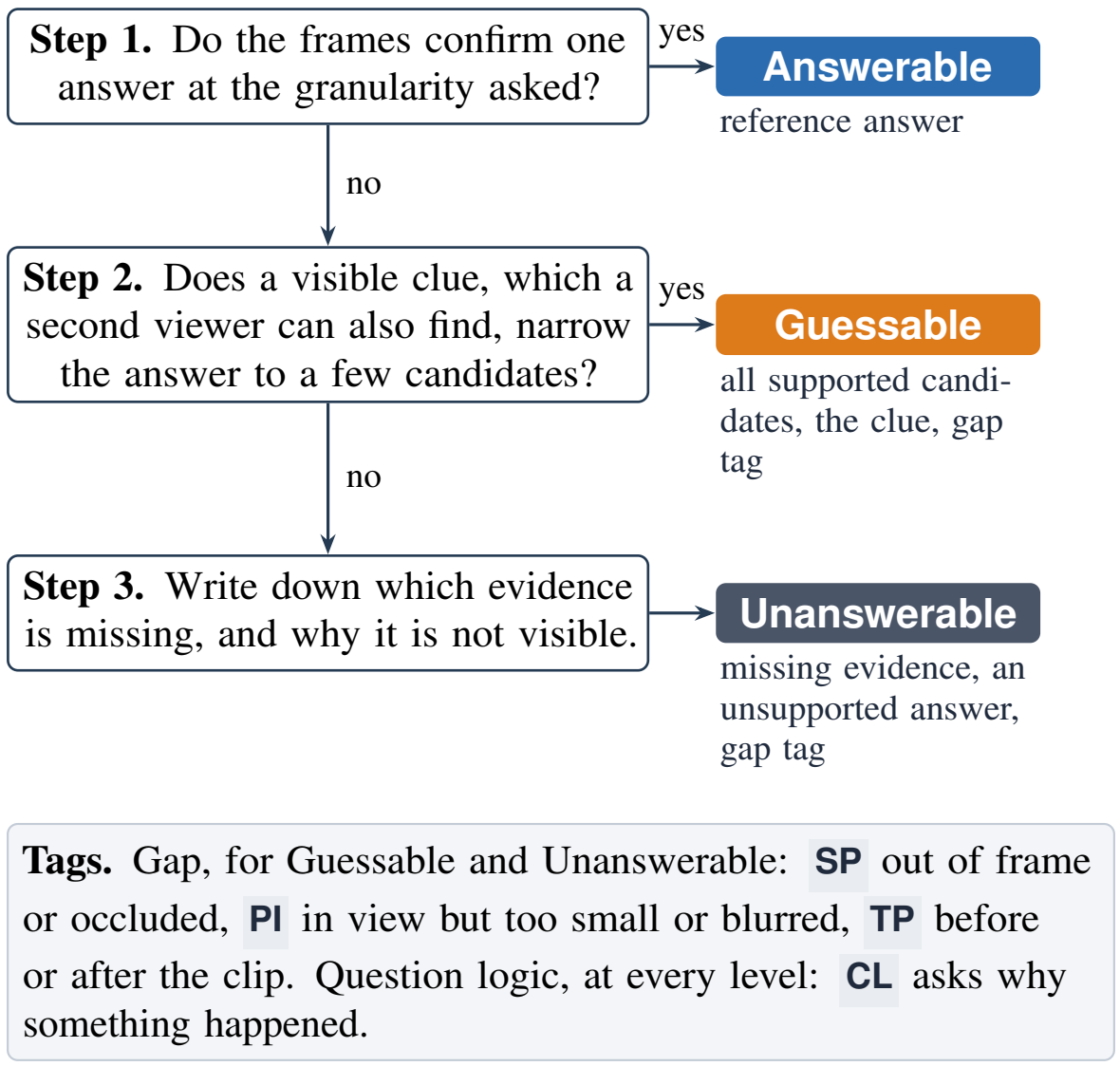

\para{Level decision.} \cref{fig:flow} shows the three steps. The second step is the one that defines the Guessable level, and two conditions keep it strict. The clue must be a concrete visible fact (an object, a state, or an action) that a second viewer can verify, which excludes guesses that rest only on what is typical. The clue must also narrow the answer to a few candidates, and the reference guess lists all of them; a response that names any listed candidate, or another answer of the same kind that agrees with the clue, counts as correct content. For Unanswerable questions the annotation records a plausible answer that the video does not support, which later serves as the wrong candidate in the perturbation set (\cref{supp:interv}). The fine-tuning set (\cref{supp:ft}) follows the same guidelines.

\para{Quality control.} Every question, reference, clue, and tag was checked against the frames a second time, and five flawed questions were removed from an initial pool. Two annotators independently labeled 60 questions and agreed on the level of 57 (95.0\%).

\subsection{Statistics}\label{supp:stats}

\begin{table}[t]
\centering
\footnotesize
\setlength{\tabcolsep}{3.1pt}
\begin{tabular}{@{}l rrr r rrr@{}}
\toprule
 & \multicolumn{3}{c}{Source} & & \multicolumn{3}{c}{Gap type} \\
\cmidrule(lr){2-4}\cmidrule(l){6-8}
Level & Charades & Ego4D & Total & CL & SP & PI & TP \\
\midrule
Answerable & 518 & 143 & 661 & 66 & -- & -- & -- \\
Guessable & 62 & 367 & 429 & 117 & 140 & 148 & 140 \\
Unanswerable & 310 & 321 & 631 & 75 & 252 & 216 & 163 \\
\midrule
Total & 890 & 831 & 1,721 & 258 & 392 & 364 & 303 \\
\bottomrule
\end{tabular}
\caption{\textbf{Statistics of \name.} Gap tags (SP: spatial, PI: perceptual, TP: temporal) apply to Guessable and Unanswerable questions; CL marks causal questions. One Guessable question has no reference guess; it is kept for response-type statistics and excluded from content metrics (G-Appr and G-Hit use 428 questions).}
\label{tab:stats}
\end{table}

\begin{figure*}[t]
\centering
\includegraphics[width=\textwidth]{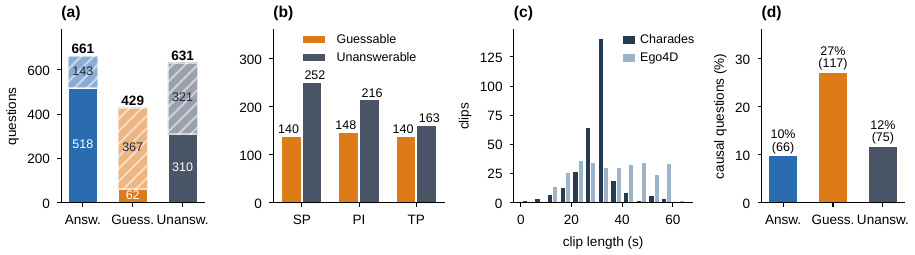}
\caption{\textbf{Composition of the main set.} (a) Questions per level and source (solid: Charades; hatched: Ego4D). (b) Gap tags of the questions with incomplete evidence. (c) Clip length. (d) Share and number of causal questions per level.}
\label{fig:stats}
\end{figure*}

\cref{tab:stats} and \cref{fig:stats} summarize the main set. Every gap tag covers at least 140 questions at each level (140, 148, and 140 Guessable questions; 252, 216, and 163 Unanswerable ones), so the per-tag rates in \cref{supp:main} rest on samples of similar size.

\para{Videos.} The 299 Charades clips last 3--59 seconds (median 30) and the 299 Ego4D clips 10--60 seconds (median 36). We use the silent versions of all clips, so no question can be answered from audio. Of the 598 videos, 529 carry three questions, 65 carry two, and 4 carry one.

\para{Why sources differ by level.} Charades clips show one scripted action in full view from a fixed camera, so most well-posed questions about them are either Answerable or, when they ask about closed containers or the time around the clip, Unanswerable. Ego4D clips show a task in progress from a moving head-mounted camera, where tools, materials, and partial views of the workspace hint at what is hidden; they supply 367 of the 429 Guessable questions. All causal Guessable and Unanswerable questions come from Ego4D, so comparisons that involve the causal flag are made within Ego4D, and \cref{supp:main} repeats the main results within each source.

\section{Prompts and Judge}
\label{supp:prompts}

\para{Prompts for the evaluated models.} Both prompts are used verbatim for every model.
\begin{promptbox}[Plain prompt (P)]
Answer the following question based only on the video.\\
Question: \{question\}\\
Give a brief and direct answer.
\end{promptbox}
\begin{promptbox}[Three-way prompt (T)]
Answer the question using only evidence visible in the video.\\
If the evidence is sufficient, answer directly.\\
If the evidence supports only a plausible guess, state the guess with clear uncertainty.\\
If the video does not provide enough evidence, answer: ``Cannot be determined from the video.''\\
Question: \{question\}
\end{promptbox}
\noindent For GPT-6.1 Terra, which receives frames, the line \emph{``The following images are frames sampled from a video in temporal order.''} is prepended. In the no-video condition the prompts are unchanged, so they still refer to a video; a model that checks its input can notice that none was given.

\begin{figure}[t]
\centering
\begin{tikzpicture}[node distance=6pt and 8pt]
\tikzset{s/.style={flowbox,font=\footnotesize,text width=2.3cm,minimum height=1.05cm},m/.style={font=\scriptsize,align=center,text=cInk,text width=2.3cm,inner sep=1pt}}
\node[headbox,font=\scriptsize\sffamily\bfseries,text width=1.25cm](r){response\\and question};
\node[s,right=of r](t){\textbf{1. Type}\\Answer, Guess, Refuse, Invalid};
\node[s,right=of t](c){\textbf{2. Content}\\matches the reference?};
\node[s,below=22pt of c](a){\textbf{3. Adoption}\\inserted content taken up?};
\node[m,below=2pt of t]{A-FR, G-FR, U-Ref, ES};
\node[m,below=2pt of c]{A-Acc, G-Appr, G-Hit};
\node[m,below=2pt of a]{candidate, premise rates};
\draw[arr](r)--(t);\draw[arr](t)--(c);\draw[arr](c.east)--++(7pt,0)|-(a.east);
\end{tikzpicture}\\[4pt]
\includegraphics[width=\linewidth]{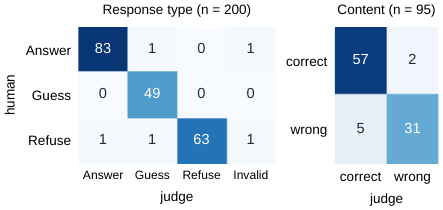}
\caption{\textbf{Scoring.} Top: the judge decides the response type first and the content second; variants with inserted content receive a third decision. The metrics that rest on each decision are listed below it. Bottom: agreement between the judge and human labels on 200 responses.}
\label{fig:judge}
\end{figure}

\para{Judge.} The judge sees only text. Its prompts are written in the language of the annotations and accept responses in either language. We summarize them here and release them verbatim with the code. The decisions are made in the order of \cref{fig:judge}.
\emph{Response type}: given the question and the response, output Answer (a definite answer without expressed uncertainty), Guess (a candidate with explicit uncertainty, \eg ``probably'', ``might'', ``looks like''), Refuse (the video cannot determine the answer and no candidate is given), or Invalid (empty or unrelated). A response that first says the answer cannot be determined and then offers a candidate is a Guess; a candidate must reach the granularity the question asks for.
\emph{Content}: for Answerable questions, the response must agree with the reference answer in its core meaning at the granularity asked (a more specific, non-contradicting answer is accepted); for Guessable questions, it must fall in the same semantic class as the reference guess and be compatible with the clue, and the reference guess is not the only acceptable one.
\emph{Adoption}: for candidate and premise variants, whether the response accepts the embedded candidate, or treats the presupposed event as having happened.

\begin{table}[t]
\centering
\scriptsize
\setlength{\tabcolsep}{3pt}
\begin{tabular}{@{}p{0.70\linewidth}l@{}}
\toprule
Response & Type \\
\midrule
``A guitar.'' & \tyA \\
``Cannot be determined from the video.'' & \tyR \\
``It looks like a guitar.'' & \tyG \\
``The object in the person's hand is a stringed instrument, likely a guitar.'' & \tyG \\
``The video does not make the reason clear; she simply stands by the stove with her arms crossed.'' & \tyR \\
``The video does not provide clear evidence of the woman's specific reason [\dots] It is plausible she is observing the person filming or waiting for an action [\dots]'' & \tyG \\
``She is waiting for the food to cook.'' & \tyA \\
\bottomrule
\end{tabular}
\caption{\textbf{Response types assigned by the judge} to model responses on the two questions of \cref{fig:wall}. A response that states doubt and still offers a candidate is a Guess; a description of the scene without a candidate is a Refuse.}
\label{tab:types}
\end{table}

\para{Calibration.} A human labeled the response type of 200 responses to 65 Answerable, 70 Guessable, and 65 Unanswerable questions, and the content of the 95 among them that answer or guess on Answerable and Guessable questions. The judge agrees with the human on 195 of the 200 response types ($\kappa=0.96$) and on 88 of the 95 content judgments ($\kappa=0.84$). \cref{fig:judge} shows where the few disagreements fall. All 49 responses that the human labeled as guesses receive the same label from the judge, so the Guess type, on which G-Appr rests, is the most reliably recognized one. The five type disagreements are atypical outputs such as ``nothing'', ``not sure'', or ``refers to a place''. On content, the judge accepts five responses that the human rejected and rejects two that the human accepted, so the two directions nearly cancel in the reported rates.

\section{Models and Inference Settings}
\label{supp:models}

\begin{table}[h]
\centering
\footnotesize
\setlength{\tabcolsep}{3pt}
\begin{tabular}{@{}lll@{}}
\toprule
Model & Visual input & Max new tokens \\
\midrule
GPT-6.1 Terra & frames, 1 fps, $\le$32 & 1,024 \\
Gemini 3.8 Flash & video file, 1 fps & 1,024 \\
Qwen3-VL-32B-Instruct & video, 2 fps, $\le$120 frames & 128$^{\ast}$ \\
Qwen3-VL-8B-Thinking & video, 2 fps, $\le$120 frames & 2,048 \\
Qwen3-VL-8B-Instruct & video, official default & 128 \\
Qwen3-VL-2B-Instruct & video, official default & 128 \\
InternVL3.5-8B & official default & 128 \\
InternVideo2.5-Chat-8B & official default & 128 \\
VideoLLaMA3-7B & 1 fps, $\le$64 frames & 128 \\
LLaVA-Video-7B-Qwen2 & 64 frames & 256 \\
\bottomrule
\end{tabular}
\caption{\textbf{Inference settings.} Open-source models decode greedily; proprietary models use API default sampling with low reasoning effort. $^{\ast}$226 three-way-prompt responses that reached the limit were regenerated with a limit of 1,024.}
\label{tab:models}
\end{table}

Every model keeps one input setting across all conditions, so each comparison in the paper (with and without the video, plain and three-way prompt, original and perturbed question) is made within a model under identical inputs. For Qwen3-VL-8B-Thinking, the text after the reasoning block is judged; responses whose reasoning does not terminate within the budget are invalid. LLaVA-Video is run with the library versions of its official release. Questions are posed in the language of the annotations under the instructions of \cref{supp:prompts}. Blank videos are 10-second black clips, and a swapped video always comes from the other source. Invalid responses (generation failures, truncated outputs, and judge-invalid outputs) are excluded from the rates of the main evaluation, and their counts are listed in \cref{tab:dist}; the fine-tuning tables count them as failures (\cref{supp:ft}).

\section{Additional Results on the Main Set}
\label{supp:main}\label{supp:es}

\subsection{All Response Types at All Levels}\label{supp:alltypes}

\begin{figure*}[t]
\centering
\includegraphics[width=\textwidth]{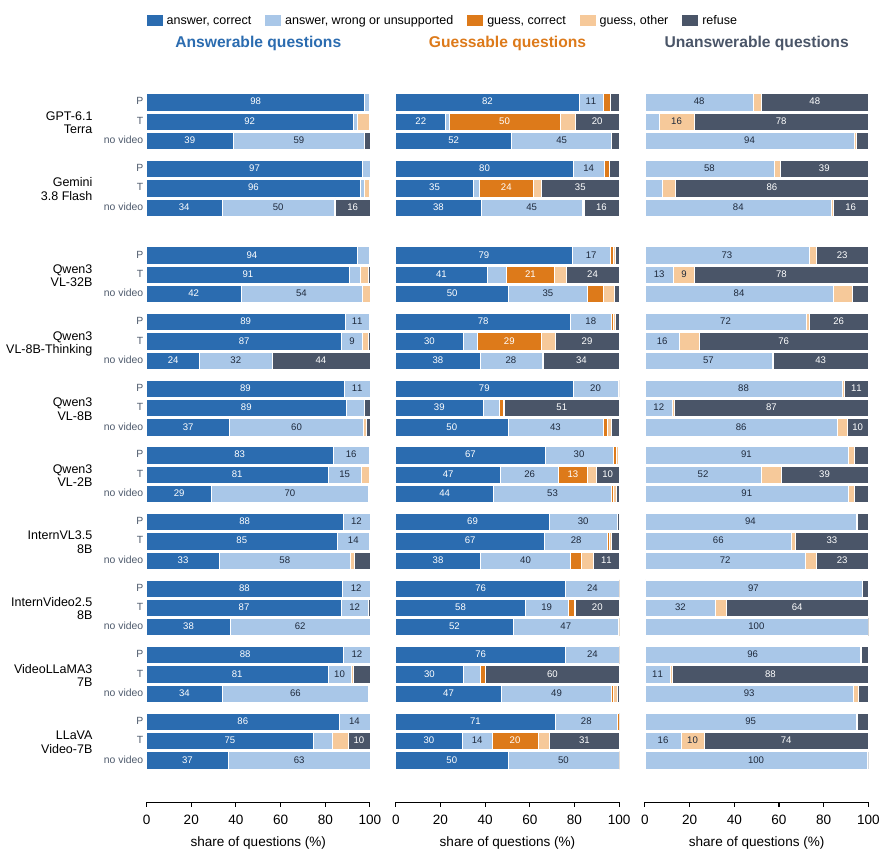}
\caption{\textbf{Responses of all ten models at every level} under the plain prompt (P), the three-way prompt (T), and the plain prompt without video. Dark orange marks the grounded guess, the response expected on Guessable questions. Numbers inside the bars are percentages; \cref{tab:dist} lists all values.}
\label{fig:distall}
\end{figure*}

\cref{fig:distall} extends Fig.~3 of the main paper, which covers Guessable questions, to all three levels and to the condition without video. Three patterns can be read directly from the figure.

\para{One response type under the plain prompt.} For most open-source models, the top bar of each group is almost entirely blue in all three panels: the model answers whether the video settles the question, only hints at the answer, or shows nothing relevant. The 32B and the reasoning Qwen3-VL models refuse about a quarter of the Unanswerable questions, and the proprietary models up to half, yet they too guess appropriately on at most 3.3\% of the Guessable ones. In the Unanswerable panel this blue is light, since every such answer is unsupported.

\para{Refusal spreads to partial evidence.} Under the three-way prompt the gray segment grows in the Unanswerable panel, as intended, and it grows in the Guessable panel as well. For VideoLLaMA3-7B and Qwen3-VL-8B it covers more than half of the Guessable questions. The correct answers that these two models gave under the plain prompt (dark blue in the Guessable panel) shrink by 46 and 40 points, so the content that the clue supported is withdrawn together with the unsupported content.

\para{Without the video, answers continue.} The bottom bar of each group shows the plain prompt without any video. The dark blue segment of the Answerable panel shrinks to 23.8--42.2\%, which confirms that correct answers depend on the video, yet the bar stays blue: the models keep answering, now mostly wrongly, and refusal remains rare for eight of the ten models.

\begin{table*}[t]
\centering
\scriptsize
\setlength{\tabcolsep}{4.2pt}
\begin{tabular}{@{}ll ccc ccc ccc cc@{}}
\toprule
 & & \multicolumn{3}{c}{Answerable} & \multicolumn{3}{c}{Guessable} & \multicolumn{3}{c}{Unanswerable} & & \\
\cmidrule(lr){3-5}\cmidrule(lr){6-8}\cmidrule(lr){9-11}
Model & Condition & Answer & Guess & Refuse & Answer & Guess & Refuse & Answer & Guess & Refuse & G-Hit & \#Invalid \\
\midrule
\multirow{3}{*}{GPT-6.1 Terra} & Plain & \underline{99.8} & 0.2 & \textbf{0.0} & 92.8 & \textbf{3.3} & 4.0 & \textbf{48.5} & 3.6 & \textbf{47.9} & \textbf{85.5} & 0 \\
 & Three-way & 94.4 & 5.3 & \underline{0.3} & 23.8 & \textbf{56.6} & 19.6 & \textbf{6.3} & 15.7 & 78.0 & \textbf{72.2} & 0 \\
 & No video & 97.7 & 0.0 & 2.3 & 96.7 & 0.0 & 3.3 & 93.7 & 1.1 & 5.2 & 51.9 & 0 \\
\midrule
\multirow{3}{*}{Gemini 3.8 Flash} & Plain & \textbf{100.0} & 0.0 & \textbf{0.0} & 93.4 & \underline{2.4} & 4.3 & \underline{57.8} & 2.8 & \underline{39.4} & \underline{81.8} & 25 \\
 & Three-way & 97.5 & 2.0 & 0.5 & 37.4 & 27.9 & 34.8 & \underline{7.6} & 6.1 & 86.2 & 59.2 & 31 \\
 & No video & 84.2 & 0.3 & 15.5 & 83.5 & 0.7 & 15.8 & 83.5 & 0.6 & 15.8 & 38.6 & 19 \\
\midrule
\multirow{3}{*}{Qwen3-VL-32B} & Plain & \underline{99.8} & 0.2 & \textbf{0.0} & 96.0 & 2.1 & 1.9 & 73.4 & 3.2 & 23.4 & 80.6 & 3 \\
 & Three-way & 95.6 & 3.5 & 0.9 & 49.7 & 26.6 & 23.8 & 12.8 & 9.4 & 77.8 & 62.1 & 0 \\
 & No video & 96.4 & 3.6 & \textbf{0.0} & 85.8 & \textbf{11.9} & 2.3 & 84.3 & 8.6 & 7.1 & \textbf{57.5} & 0 \\
\midrule
\multirow{3}{*}{Qwen3-VL-8B-Thinking} & Plain & \underline{99.8} & 0.2 & \textbf{0.0} & 96.5 & 1.9 & 1.7 & 72.2 & 1.3 & 26.5 & 79.4 & 25 \\
 & Three-way & 96.5 & 2.7 & 0.8 & 36.4 & \underline{34.8} & 28.7 & 15.5 & 8.7 & 75.8 & 59.0 & 14 \\
 & No video & 56.1 & 0.0 & 43.9 & 65.5 & 0.7 & 33.8 & \textbf{56.9} & 0.5 & \textbf{42.6} & 38.2 & 41 \\
\midrule
\multirow{3}{*}{Qwen3-VL-8B} & Plain & \textbf{100.0} & 0.0 & \textbf{0.0} & 99.5 & 0.2 & \underline{0.2} & 88.2 & 1.1 & 10.6 & 79.2 & 13 \\
 & Three-way & 97.3 & 0.0 & 2.7 & 46.4 & 2.3 & 51.3 & 12.4 & 0.8 & \underline{86.8} & 40.7 & 0 \\
 & No video & 97.1 & 1.2 & 1.7 & 92.8 & 3.7 & 3.5 & 86.3 & 4.2 & 9.6 & 52.3 & 5 \\
\midrule
\multirow{3}{*}{Qwen3-VL-2B} & Plain & \underline{99.8} & 0.0 & \underline{0.2} & 97.2 & 2.3 & 0.5 & 90.9 & 3.0 & 6.1 & 68.5 & 10 \\
 & Three-way & 96.2 & 3.3 & 0.5 & 72.9 & 16.8 & \underline{10.3} & 52.1 & 8.8 & 39.2 & 59.8 & 4 \\
 & No video & 99.2 & 0.3 & 0.5 & 96.6 & 2.2 & 1.2 & 91.2 & 2.6 & 6.2 & 44.8 & 22 \\
\midrule
\multirow{3}{*}{InternVL3.5-8B} & Plain & \textbf{100.0} & 0.0 & \textbf{0.0} & 99.1 & 0.2 & 0.7 & 94.4 & 0.6 & 5.0 & 68.8 & 9 \\
 & Three-way & \textbf{99.8} & 0.0 & \textbf{0.2} & 94.8 & 1.9 & \textbf{3.3} & 65.7 & 1.7 & 32.6 & \underline{67.6} & 4 \\
 & No video & 91.1 & 2.1 & 6.8 & 78.3 & \underline{10.3} & 11.4 & \underline{71.7} & 4.9 & \underline{23.3} & 42.6 & 2 \\
\midrule
\multirow{3}{*}{InternVideo2.5-8B} & Plain & \textbf{100.0} & 0.0 & \textbf{0.0} & 100.0 & 0.0 & \textbf{0.0} & 97.1 & 0.0 & 2.9 & 76.1 & 6 \\
 & Three-way & \underline{99.1} & 0.0 & 0.9 & 77.2 & 3.3 & 19.5 & 31.7 & 4.6 & 63.7 & 60.6 & 4 \\
 & No video & \textbf{100.0} & 0.0 & \textbf{0.0} & 99.8 & 0.0 & \underline{0.2} & 99.8 & 0.0 & 0.2 & \underline{52.5} & 8 \\
\midrule
\multirow{3}{*}{VideoLLaMA3-7B} & Plain & \textbf{100.0} & 0.0 & \textbf{0.0} & 100.0 & 0.0 & \textbf{0.0} & 96.5 & 0.3 & 3.2 & 76.1 & 6 \\
 & Three-way & 91.6 & 0.8 & 7.6 & 38.0 & 2.3 & 59.7 & 11.3 & 1.0 & \textbf{87.8} & 32.0 & 4 \\
 & No video & \underline{99.4} & 0.3 & \underline{0.3} & 96.7 & 2.6 & 0.7 & 93.3 & 2.1 & 4.6 & 47.8 & 7 \\
\midrule
\multirow{3}{*}{LLaVA-Video-7B} & Plain & \textbf{100.0} & 0.0 & \textbf{0.0} & 99.3 & 0.7 & \textbf{0.0} & 94.6 & 0.3 & 5.1 & 72.0 & 6 \\
 & Three-way & 83.2 & 7.0 & 9.8 & 43.2 & 25.7 & 31.1 & 16.2 & 10.2 & 73.6 & 50.1 & 4 \\
 & No video & \textbf{100.0} & 0.0 & \textbf{0.0} & 100.0 & 0.0 & \textbf{0.0} & 99.5 & 0.2 & 0.3 & 50.2 & 2 \\
\bottomrule
\end{tabular}
\caption{\textbf{Response-type distribution (\%)} per level under the plain prompt, the three-way prompt, and the plain prompt without video. G-Hit: Guessable questions whose content matches the reference guess under an answer or a guess. \#Invalid: excluded responses out of 1,721.}
\label{tab:dist}
\end{table*}

\subsection{Results within Each Video Source}\label{supp:bysource}

\begin{figure*}[t]
\centering
\includegraphics[width=\textwidth]{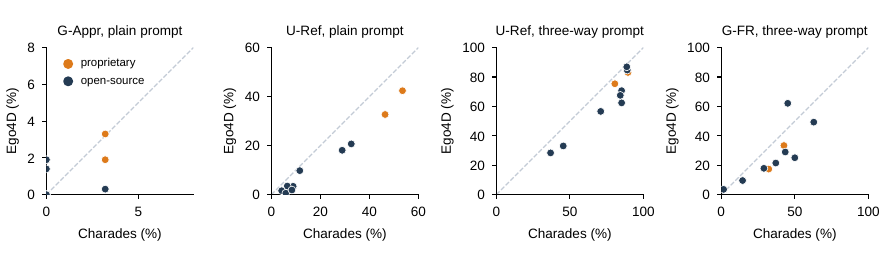}
\caption{\textbf{The same pattern in both video sources.} Each dot is one model; its position gives a rate on Charades questions (horizontal) and on Ego4D questions (vertical). Dots near the diagonal mean that a model behaves alike on both sources. Charades has 62 Guessable questions, so its Guessable rates are coarse.}
\label{fig:source}
\end{figure*}

\begin{table*}[t]
\centering
\scriptsize
\setlength{\tabcolsep}{4.2pt}
\begin{tabular}{@{}l ccc ccc ccc ccc@{}}
\toprule
 & \multicolumn{6}{c}{Charades} & \multicolumn{6}{c}{Ego4D} \\
\cmidrule(lr){2-7}\cmidrule(l){8-13}
 & \multicolumn{3}{c}{Plain} & \multicolumn{3}{c}{Three-way} & \multicolumn{3}{c}{Plain} & \multicolumn{3}{c}{Three-way} \\
\cmidrule(lr){2-4}\cmidrule(lr){5-7}\cmidrule(lr){8-10}\cmidrule(l){11-13}
Model & A-Acc & G-Appr & U-Ref & A-Acc & G-Appr & U-Ref & A-Acc & G-Appr & U-Ref & A-Acc & G-Appr & U-Ref \\
\midrule
GPT-6.1 Terra & \textbf{98.1} & \textbf{3.2} & \textbf{53.5} & \underline{94.4} & \textbf{35.5} & 80.6 & \textbf{96.5} & \textbf{3.3} & \textbf{42.4} & 85.3 & \textbf{52.2} & 75.4 \\
Gemini 3.8 Flash & \underline{97.3} & \textbf{3.2} & \underline{46.4} & \textbf{95.5} & \underline{21.3} & \textbf{89.5} & \underline{94.4} & \underline{1.9} & \underline{32.7} & \textbf{96.4} & 24.9 & 83.1 \\
Qwen3-VL-32B & 95.0 & \underline{0.0} & 28.9 & 90.7 & 21.0 & 85.2 & 92.3 & \underline{1.9} & 18.1 & \underline{90.9} & 21.3 & 70.7 \\
Qwen3-VL-8B-Thinking & 89.2 & \underline{0.0} & 32.6 & 88.8 & 12.9 & 84.3 & 87.4 & 1.4 & 20.7 & 82.5 & \underline{31.5} & 67.5 \\
Qwen3-VL-8B & 90.1 & \underline{0.0} & 11.6 & 89.4 & 0.0 & \underline{89.0} & 83.2 & 0.0 & 9.8 & 88.8 & 1.9 & \underline{84.7} \\
Qwen3-VL-2B & 84.6 & \underline{0.0} & 8.9 & 83.2 & 4.8 & 45.5 & 79.0 & \underline{1.9} & 3.4 & 74.1 & 14.2 & 33.1 \\
InternVL3.5-8B & 90.3 & \underline{0.0} & 6.5 & 87.5 & 0.0 & 36.9 & 79.0 & 0.0 & 3.5 & 77.6 & 1.1 & 28.4 \\
InternVideo2.5-8B & 90.2 & \underline{0.0} & 4.2 & 88.0 & 4.8 & 71.0 & 79.0 & 0.0 & 1.6 & 84.5 & 2.2 & 56.6 \\
VideoLLaMA3-7B & 89.9 & \underline{0.0} & 5.9 & 85.5 & 0.0 & 88.7 & 82.5 & 0.0 & 0.6 & 65.2 & 2.2 & \textbf{86.9} \\
LLaVA-Video-7B & 87.2 & \textbf{3.2} & 8.4 & 76.8 & 14.5 & 85.2 & 82.5 & 0.3 & 1.9 & 67.1 & 21.4 & 62.4 \\
\bottomrule
\end{tabular}
\caption{\textbf{Main results by video source (\%).} Under the three-way prompt, U-Ref rises by 25 to 86 points in both sources, and G-Appr stays below 36\% for every model except GPT-6.1 Terra on Ego4D.}
\label{tab:source}
\end{table*}

The two sources differ in viewpoint and in their mix of levels (\cref{supp:data}), so we repeat the main analysis within each source. \cref{fig:source} and \cref{tab:source} show that the diagnosis does not depend on the source. Under the plain prompt, appropriate guessing stays at or below 3.3\% in both sources for every model. Across the ten models, U-Ref on Charades and U-Ref on Ego4D correlate at $r=0.99$ under the plain prompt and $r=0.95$ under the three-way prompt, and G-FR under the three-way prompt at $r=0.79$. A model's way of responding is thus a property of the model and the prompt, and it carries over from third-person scripted clips to first-person recordings.

\subsection{Vision-Necessary Subset}\label{supp:vn}

\begin{table}[t]
\centering
\scriptsize
\setlength{\tabcolsep}{3.2pt}
\begin{tabular}{@{}l cc ccc c@{}}
\toprule
 & \multicolumn{2}{c}{Plain} & \multicolumn{3}{c}{Three-way} & No video \\
\cmidrule(lr){2-3}\cmidrule(lr){4-6}\cmidrule(l){7-7}
Model & All & VN & All & VN & VN A-FR & All \\
\midrule
GPT-6.1 Terra & \textbf{97.7} & \textbf{97.2} & \underline{92.4} & \underline{89.9} & \textbf{0.0} & 38.7 \\
Gemini 3.8 Flash & \underline{96.7} & \underline{93.8} & \textbf{95.7} & \textbf{93.2} & \underline{0.6} & 33.7 \\
Qwen3-VL-32B & 94.4 & 87.6 & 90.8 & 86.5 & 1.1 & 42.2 \\
Qwen3-VL-8B-Thinking & 88.8 & 80.3 & 87.4 & 81.4 & 1.1 & 23.8 \\
Qwen3-VL-8B & 88.6 & 78.7 & 89.3 & 81.5 & 3.9 & 37.2 \\
Qwen3-VL-2B & 83.4 & 75.3 & 81.2 & 75.3 & \underline{0.6} & 29.2 \\
InternVL3.5-8B & 87.9 & 81.9 & 85.3 & 77.0 & \underline{0.6} & 32.7 \\
InternVideo2.5-8B & 87.7 & 78.1 & 87.3 & 77.0 & 1.7 & 37.6 \\
VideoLLaMA3-7B & 88.3 & 75.7 & 81.2 & 70.8 & 11.2 & 33.9 \\
LLaVA-Video-7B & 86.2 & 76.8 & 74.7 & 65.5 & 13.6 & 36.6 \\
\bottomrule
\end{tabular}
\caption{\textbf{A-Acc on the vision-necessary subset (\%).} VN: the 178 Answerable questions that none of the ten models answers correctly without the video. VN A-FR: false refusals on this subset.}
\label{tab:vn}
\end{table}

483 of the 661 Answerable questions are answered correctly without the video by at least one of the ten models, typically because the question narrows the answer to a common object or action. We keep them, since responding from priors is part of the behavior we measure, and repeat the analysis on the remaining 178 (\cref{tab:vn}). A-Acc stays between 75.3\% and 97.2\% under the plain prompt, so the perception finding of \cref{sec:see} holds on questions that language priors cannot answer.

\subsection{Bound on ES without Guessing}\label{supp:esbound}

For fixed level frequencies, $I(L;R)$ is convex in the response policy $p(R\mid L)$. Over policies that use only the answer and refusal types, its maximum is therefore reached at a deterministic policy, which maps each level to one type. The best such policy separates Answerable questions from the other two levels and gives $\text{ES}=100\cdot H_2(661/1{,}721)/H(L)=100\cdot 0.961/1.561=61.6$, where $H_2$ is the binary entropy. Exceeding this value requires a third response type, the guess. GPT-6.1 Terra, which guesses appropriately on about half of the Guessable questions, reaches the highest ES among the ten evaluated models (44.9), and guess-aware tuning with both prompts raises Qwen3-VL-8B from 31.1 to 48.5. The distance to the bound shows how much information a response type can still carry once the guess level is used consistently.

\subsection{Gap Tags and Causal Questions}\label{supp:gapcl}

\begin{figure}[t]
\centering
\includegraphics[width=\linewidth]{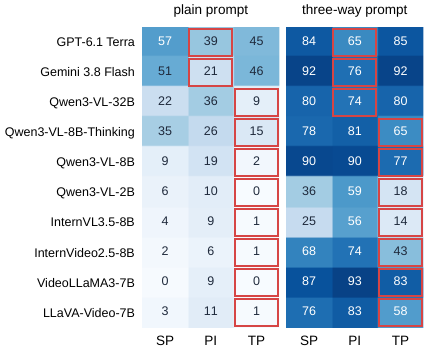}
\caption{\textbf{Refusal of Unanswerable questions by gap tag (U-Ref, \%).} The red frame marks the tag that each model refuses least. Under the plain prompt this is the temporal gap for every open-source model; under the three-way prompt, for seven of the eight.}
\label{fig:gapmap}
\end{figure}

\begin{table*}[t]
\centering
\scriptsize
\setlength{\tabcolsep}{4.4pt}
\begin{tabular}{@{}l ccc ccc ccc ccc@{}}
\toprule
 & \multicolumn{3}{c}{U-Ref, plain} & \multicolumn{3}{c}{U-Ref, three-way} & \multicolumn{3}{c}{G-Appr, three-way} & \multicolumn{3}{c}{G-FR, three-way} \\
\cmidrule(lr){2-4}\cmidrule(lr){5-7}\cmidrule(lr){8-10}\cmidrule(l){11-13}
Model & SP & PI & TP & SP & PI & TP & SP & PI & TP & SP & PI & TP \\
\midrule
GPT-6.1 Terra & \textbf{57.1} & \textbf{39.4} & \underline{44.8} & 84.1 & 65.3 & \underline{85.3} & \textbf{45.7} & \textbf{60.1} & \textbf{42.9} & 27.1 & \textbf{6.1} & 26.4 \\
Gemini 3.8 Flash & \underline{50.8} & 20.8 & \textbf{45.9} & \textbf{91.9} & 75.5 & \textbf{91.8} & \underline{28.1} & 23.3 & 21.7 & 33.3 & 15.8 & 56.5 \\
Qwen3-VL-32B & 21.9 & \underline{35.6} & 9.3 & 79.8 & 74.1 & 79.8 & 14.3 & \underline{29.1} & 20.0 & 26.4 & 15.5 & 30.0 \\
Qwen3-VL-8B-Thinking & 34.6 & 25.7 & 14.7 & 78.4 & 81.3 & 64.6 & 21.6 & 28.4 & \underline{36.4} & 32.4 & 27.0 & 27.1 \\
Qwen3-VL-8B & 9.2 & 18.7 & 2.5 & \underline{90.5} & \underline{89.8} & 77.3 & 0.7 & 3.4 & 0.7 & 50.7 & 50.7 & 52.9 \\
Qwen3-VL-2B & 6.4 & 10.4 & 0.0 & 35.9 & 59.1 & 17.9 & 11.4 & 12.8 & 14.3 & \underline{13.6} & 11.5 & \underline{5.7} \\
InternVL3.5-8B & 4.0 & 9.4 & 0.6 & 24.7 & 56.0 & 13.6 & 0.7 & 0.7 & 1.4 & \textbf{0.0} & \underline{8.8} & \textbf{0.7} \\
InternVideo2.5-8B & 1.6 & 6.0 & 0.6 & 67.7 & 74.5 & 42.9 & 1.4 & 4.8 & 1.4 & 20.1 & 19.7 & 18.6 \\
VideoLLaMA3-7B & 0.4 & 8.9 & 0.0 & 86.9 & \textbf{92.6} & 82.8 & 2.1 & 2.0 & 1.4 & 55.0 & 57.4 & 66.4 \\
LLaVA-Video-7B & 2.8 & 11.3 & 0.6 & 75.8 & 82.9 & 57.8 & 15.8 & 25.7 & 19.3 & 30.9 & 33.8 & 28.6 \\
\bottomrule
\end{tabular}
\caption{\textbf{Results by gap type (\%).} The U-Ref columns give the values of \cref{fig:gapmap}; the remaining columns split G-Appr and G-FR under the three-way prompt.}
\label{tab:gap}
\end{table*}

\para{Gap types.} A perceptual gap leaves a small or blurred target in view, and a spatial gap often leaves part of the target or the object that hides it. A temporal gap leaves no trace in the frames: nothing in the clip shows that something happened before it. \cref{fig:gapmap} and \cref{tab:gap} show that open-source models notice missing evidence mainly when the gap is visible. InternVL3.5-8B refuses 56.0\% of perceptual, 24.7\% of spatial, and 13.6\% of temporal Unanswerable questions under the three-way prompt, and Qwen3-VL-2B 59.1\%, 35.9\%, and 17.9\%. The gap type that only video can pose is thus the one where these models most often answer what they cannot see. The two proprietary models refuse temporal gaps about as often as spatial ones, which suggests that awareness of the clip boundaries can be acquired and is not yet common.

\begin{table}[t]
\centering
\scriptsize
\setlength{\tabcolsep}{3.4pt}
\begin{tabular}{@{}l cc cc cc@{}}
\toprule
 & \multicolumn{2}{c}{G: guess} & \multicolumn{2}{c}{U: guess} & \multicolumn{2}{c}{U: refuse} \\
\cmidrule(lr){2-3}\cmidrule(lr){4-5}\cmidrule(l){6-7}
Model & CL & non-CL & CL & non-CL & CL & non-CL \\
\midrule
GPT-6.1 Terra & \textbf{68.4} & \textbf{54.8} & 33.3 & 12.6 & 66.7 & 78.0 \\
Gemini 3.8 Flash & 35.4 & 24.8 & 9.6 & 5.8 & \textbf{80.8} & 83.8 \\
Qwen3-VL-32B & 23.9 & 27.6 & 30.7 & 6.1 & 36.0 & 81.3 \\
Qwen3-VL-8B-Thinking & 56.4 & \underline{28.9} & 54.1 & 2.5 & 27.0 & 80.0 \\
Qwen3-VL-8B & 7.7 & 0.4 & \textbf{5.3} & \textbf{0.0} & 70.7 & \underline{89.0} \\
Qwen3-VL-2B & 48.3 & 4.0 & 52.0 & 2.4 & 10.7 & 40.0 \\
InternVL3.5-8B & 7.0 & 0.0 & 10.8 & \underline{0.4} & 1.4 & 36.6 \\
InternVideo2.5-8B & 4.4 & 2.4 & \underline{6.8} & 2.4 & 18.9 & 67.9 \\
VideoLLaMA3-7B & 6.0 & 1.2 & 8.1 & \textbf{0.0} & \underline{75.7} & \textbf{90.2} \\
LLaVA-Video-7B & \underline{60.7} & 11.6 & 65.3 & 3.7 & 16.0 & 76.6 \\
\bottomrule
\end{tabular}
\caption{\textbf{Causal versus other questions} (Ego4D, three-way prompt, \%). Guess: share of questions met with a guess, regardless of content. Ego4D has 117 causal and 250 other Guessable questions, and 75 causal and 246 other Unanswerable ones.}
\label{tab:causal}
\end{table}

\para{Causal questions.} \cref{tab:causal} compares questions that ask for a cause with the others, within Ego4D. For nine of the ten models, guesses are more frequent on causal than on other Guessable questions, and for every model they are more frequent on causal than on other Unanswerable questions. For six of the eight open-source models, the guess rate on causal Unanswerable questions, which have no clue, is at least as high as on causal Guessable ones; the two proprietary models guess clearly less often when the clue is absent (33.3\% versus 68.4\% for GPT-6.1 Terra), so their guesses are partly tied to the evidence. Every model also refuses causal Unanswerable questions less often than other Unanswerable ones, by 3 to 61 points. The word \emph{why} thus acts as a license to speculate that is largely independent of the evidence; \cref{fig:wall} shows a typical case.

\section{Full Perturbation Results}
\label{supp:interv}

\subsection{Construction of the Perturbation Set}\label{supp:construct}

\begin{figure*}[t]
\centering
{\footnotesize\setlength{\tabcolsep}{4pt}
\begin{tabular}{@{}llp{0.33\textwidth}lr@{}}
\toprule
Group & Perturbation & What changes & Expected response & Items \\
\midrule
\multirow{5}{*}{Question text, Answerable base} & Two candidates & the correct and a wrong candidate are offered & the correct candidate & 200 \\
 & Correct candidate & ``\dots Is it X?'' with the reference answer X & yes & 200 \\
 & Wrong candidate & ``\dots Is it Y?'' with a distractor Y that the frames rule out & no, it is X & 200 \\
 & Out of scope & the question moves to the time before or after the clip, or outside the frame & refuse & 200 \\
 & False premise & an event that never happens is presupposed & refuse or reject the premise & 200 \\
\midrule
\multirow{2}{*}{Question text, Unanswerable base} & Wrong candidate & ``\dots Is it Y?'' with a plausible unsupported answer Y & refuse & 100 \\
 & Visibility claim & a sentence states that the video shows the answer clearly & refuse & 100 \\
\midrule
\multirow{3}{*}{Video, Answerable base} & Removed & the question is asked without any video & refuse & all \\
 & Blank & a 10-second black clip replaces the video & refuse & 200 \\
 & Swapped & an unrelated clip from the other source replaces the video & refuse & 197 \\
\midrule
Question text, A and G base & Redundant qualifiers & one to four true but unneeded details are added & unchanged & 198 \\
\bottomrule
\end{tabular}}\\[8pt]
\footnotesize
\setlength{\tabcolsep}{4pt}
\renewcommand{\arraystretch}{1.25}
\newcommand{\ins}[1]{{\setlength{\fboxsep}{1pt}\colorbox{cAl}{#1}}}
\newcommand{\thumb}[1]{\raisebox{-.42\height}{\includegraphics[height=0.95cm]{#1}}}
\begin{minipage}[c]{0.155\textwidth}\centering
\begin{tikzpicture}[inner sep=0]\node(a){\includegraphics[width=\linewidth]{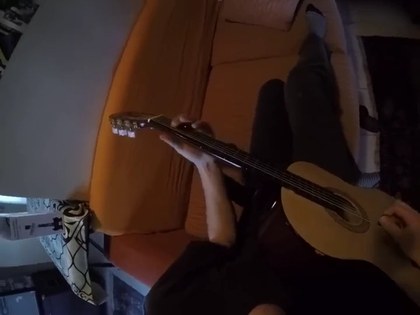}};\node[anchor=south west,fill=black,fill opacity=.55,text opacity=1,text=white,font=\tiny\sffamily,inner sep=1.2pt] at (a.south west){3 s};\end{tikzpicture}\\[2pt]\begin{tikzpicture}[inner sep=0]\node(a){\includegraphics[width=\linewidth]{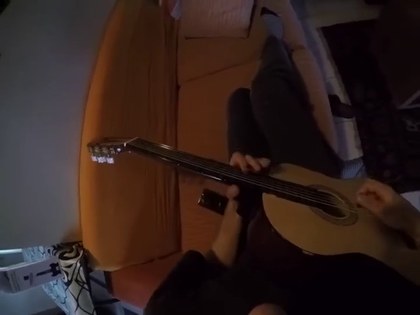}};\node[anchor=south west,fill=black,fill opacity=.55,text opacity=1,text=white,font=\tiny\sffamily,inner sep=1.2pt] at (a.south west){7 s};\end{tikzpicture}\\[2pt]
{\sffamily\bfseries\color{cNavy} video fixed,\\ question changes}
\end{minipage}\hfill
\begin{minipage}[c]{0.83\textwidth}
\begin{tabular}{@{}p{0.47\linewidth}p{0.20\linewidth}p{0.27\linewidth}@{}}
\toprule
\textbf{Question} (inserted words highlighted) & \textbf{Plain prompt} & \textbf{Three-way prompt} \\
\midrule
What type of instrument is the wearer playing? & \tyA\ ``guitar'' \yes & \tyA\ ``guitar'' \yes \\
\ins{Choose between an accordion and a guitar:} what type of \dots? & \tyA\ ``guitar'' \yes & \tyA\ ``guitar'' \yes \\
What type of instrument \dots? \ins{Is it a guitar?} & \tyA\ ``Yes.'' \yes & \tyA\ ``Yes, it is a guitar.'' \yes \\
What type of instrument \dots? \ins{Is it an accordion?} & \tyA\ ``No.'' \yes & \tyA\ ``No, it is a guitar.'' \yes \\
\ins{After finishing the piece, where does the wearer put} \ins{the instrument back?} & \tyA\ ``on the couch'' \no & \tyR\ ``Cannot be determined.'' \yes \\
\ins{After the lamp next to the sofa suddenly goes out,} what type of instrument does the wearer keep playing? & \tyA\ ``guitar'' \no & \tyA\ ``guitar'' \no \\
\bottomrule
\end{tabular}
\end{minipage}\\[6pt]
\begin{minipage}[c]{0.155\textwidth}\centering
{\sffamily\bfseries\color{cNavy} question fixed,\\ video changes}
\end{minipage}\hfill
\begin{minipage}[c]{0.83\textwidth}
\begin{tabular}{@{}p{0.20\linewidth}p{0.255\linewidth}p{0.20\linewidth}p{0.27\linewidth}@{}}
\toprule
\textbf{Video} & \textbf{Question} & \textbf{Plain prompt} & \textbf{Three-way prompt} \\
\midrule
\emph{removed} & \multirow{3}{=}[-14pt]{What type of instrument is the wearer playing?} & \tyA\ ``guitar'' \no & \tyR\ ``Cannot be determined.'' \yes \\
\raisebox{-.42\height}{\includegraphics[height=0.95cm]{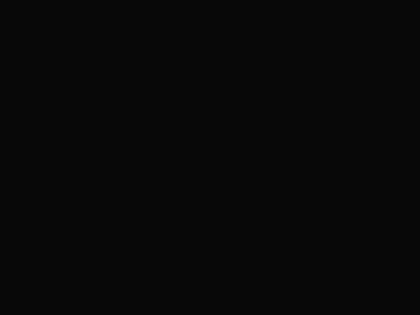}}\ \emph{blank} & & \tyA\ ``guitar'' \no & \tyR\ ``Cannot be determined.'' \yes \\[10pt]
\raisebox{-.42\height}{\includegraphics[height=0.95cm]{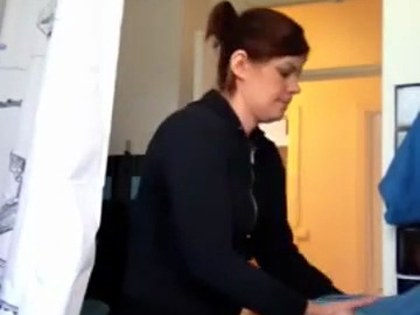}}\ \emph{swapped} & & \tyA\ ``guitar'' \no & \tyR\ ``Cannot be determined.'' \yes \\[6pt]
\bottomrule
\end{tabular}
\end{minipage}
\caption{\textbf{The perturbation set.} Top: the perturbation families. Each changes one input of a base question and leaves the other fixed; with the 300 unchanged base questions, the first three groups give 1,897 variants, and every variant is asked under both prompts. Bottom: one Answerable question under every perturbation, with the responses of Qwen3-VL-8B. With the video fixed, the model confirms the correct candidate and rejects the wrong one, so it checks suggested content against the video; under the plain prompt it also answers a question about the time after the clip, and under both prompts it answers the question with the false premise without objecting to it. With the question fixed, the answer under the plain prompt stays ``guitar'' whether the video is removed, blank, or swapped, and the three-way prompt turns all three into refusals. \yes/\no: the response matches or misses the expected response.}
\label{fig:family}
\end{figure*}

\cref{fig:family} lists the perturbation families and shows all of them on one base question.

\para{Base questions.} The 200 Answerable base questions are drawn in two equal strata, questions that most models answered correctly in a pilot run and questions that most answered wrongly, with at most two per video. Using both strata lets the set measure gains from a correct candidate as well as losses from a wrong one. The 100 Unanswerable base questions are balanced over the gap tags (34 spatial, 33 perceptual, 33 temporal), with one per video. The unchanged base question is re-asked in the same run, and every effect is measured against this re-asked version.

\para{Wording.} Candidate variants use four fixed phrasings in equal shares (\eg ``\dots Is it X?'', ``\dots It should be X, right?''), and the correct and the wrong candidate of an item always share one phrasing, so their difference cannot come from the wording. The distractor of an Answerable question is an answer that the frames rule out, in most cases one that a model actually produced. Visibility claims use four phrasings such as ``The video shows it clearly: \dots''. Out-of-scope and false-premise variants are written per question after inspecting the frames: 148 out-of-scope variants ask about the time before or after the clip and 52 about hidden or off-screen content, and the presupposed events cover object or scene changes (90), actions of the main person (77), and an extra animal or person (33).

\para{Videos.} A swapped video comes from the other source and shows a different kind of scene, so that the question clearly does not fit it; three questions that any video could answer have no swapped version. The blank clip keeps the input format of a normal video, which separates ``no visual content'' from ``no video input''.

\subsection{Results}\label{supp:pres}

\begin{table}[t]
\centering
\scriptsize
\setlength{\tabcolsep}{2.8pt}
\resizebox{\linewidth}{!}{%
\begin{tabular}{@{}l cc cc cc@{}}
\toprule
 & \multicolumn{2}{c}{Qwen3-VL-2B} & \multicolumn{2}{c}{Qwen3-VL-8B} & \multicolumn{2}{c}{VideoLLaMA3} \\
\cmidrule(lr){2-3}\cmidrule(lr){4-5}\cmidrule(l){6-7}
Variant: metric & P & T & P & T & P & T \\
\midrule
\multicolumn{7}{@{}l}{\textit{Answerable base questions (200), video present}}\\
Original: A-Acc $\uparrow$ & 88.0 & 83.0 & \textbf{92.0} & \textbf{95.0} & \underline{88.5} & \underline{86.0} \\
Two candidates: A-Acc $\uparrow$ & \underline{91.0} & \underline{92.0} & \textbf{95.0} & \textbf{97.5} & 89.0 & 89.5 \\
Correct candidate: adopted $\uparrow$ & \underline{95.5} & \underline{98.5} & 94.5 & \textbf{99.0} & \textbf{97.0} & 95.5 \\
Wrong candidate: adopted $\downarrow$ & 31.0 & 39.5 & \textbf{5.5} & \textbf{30.5} & \underline{29.6} & \underline{38.5} \\
False premise: accepted $\downarrow$ & 37.5 & 71.0 & \textbf{19.1} & \textbf{26.0} & \underline{31.0} & \underline{29.1} \\
False premise: refused $\uparrow$ & \textbf{0.0} & 3.5 & \textbf{0.0} & \textbf{26.0} & \textbf{0.0} & \underline{23.6} \\
Out of scope: refused $\uparrow$ & \underline{1.5} & 26.9 & \textbf{5.6} & \textbf{81.5} & 0.0 & \underline{74.5} \\
\midrule
\multicolumn{7}{@{}l}{\textit{Answerable base questions, video removed}}\\
Original: refused $\uparrow$ & 0.0 & \textbf{100.0} & \textbf{1.5} & 87.0 & \underline{0.5} & \underline{98.5} \\
Original: A-Acc $\downarrow$ & \underline{36.0} & \textbf{0.0} & 43.5 & 6.0 & \textbf{33.0} & \underline{1.0} \\
Two candidates: A-Acc $\downarrow$ & \underline{52.3} & \textbf{10.0} & \textbf{49.0} & 39.5 & 53.5 & \underline{16.5} \\
Correct candidate: adopted $\downarrow$ & 82.0 & \textbf{2.5} & \textbf{50.0} & 33.0 & \underline{67.5} & \underline{10.0} \\
Wrong candidate: adopted $\downarrow$ & 77.5 & \textbf{1.0} & \textbf{39.5} & 26.5 & \underline{59.0} & \underline{8.0} \\
Out of scope: refused $\uparrow$ & \underline{2.0} & \textbf{100.0} & \textbf{2.5} & \underline{97.5} & 1.0 & \textbf{100.0} \\
False premise: accepted $\downarrow$ & 46.5 & \textbf{0.0} & \textbf{39.5} & \underline{5.5} & \underline{41.7} & \textbf{0.0} \\
\midrule
\multicolumn{7}{@{}l}{\textit{Answerable base questions, video replaced}}\\
Blank video: refused $\uparrow$ & \underline{0.0} & 55.1 & \textbf{2.5} & \textbf{99.5} & \underline{0.0} & \underline{85.4} \\
Swapped video: refused $\uparrow$ & \underline{0.5} & 30.6 & \textbf{20.3} & \textbf{95.9} & 0.0 & \underline{88.8} \\
\midrule
\multicolumn{7}{@{}l}{\textit{Unanswerable base questions (100)}}\\
Original: refused $\uparrow$ & \underline{5.0} & 36.0 & \textbf{12.1} & \underline{81.0} & 3.0 & \textbf{85.0} \\
Visibility claim: refused $\uparrow$ & \underline{13.1} & 38.0 & \textbf{13.3} & \underline{78.0} & 7.1 & \textbf{83.0} \\
Wrong candidate: refused $\uparrow$ & \textbf{2.0} & 30.0 & \underline{0.0} & \underline{66.0} & \underline{0.0} & \textbf{78.0} \\
Wrong candidate: adopted $\downarrow$ & 73.0 & 58.0 & \textbf{19.0} & \underline{30.0} & \underline{70.7} & \textbf{22.0} \\
Original, no video: refused $\uparrow$ & 3.1 & \textbf{100.0} & \textbf{9.0} & \textbf{100.0} & \underline{4.1} & \textbf{100.0} \\
Wrong candidate, no video: adopted $\downarrow$ & 75.0 & \textbf{0.0} & \textbf{31.0} & 5.0 & \underline{63.0} & \underline{2.0} \\
\bottomrule
\end{tabular}}
\caption{\textbf{All perturbation conditions for the open-source models (\%).} The proprietary models are evaluated on the conditions with a present, blank, or swapped video (Tab.~3 of the main paper). P: plain prompt; T: three-way prompt. Adopted: the response accepts the embedded candidate. Accepted: the response treats the presupposed event as real.}
\label{tab:induce}
\end{table}

\cref{tab:induce} lists every condition of the perturbation set for the three models that run it in full.

\begin{figure}[t]
\centering
\includegraphics[width=\linewidth]{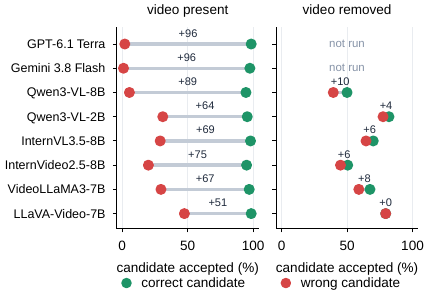}
\caption{\textbf{Candidate check on eight models} (plain prompt). Each row shows how often a model accepts an embedded correct candidate (green) and a wrong one (red); the number is their difference in points.}
\label{fig:cand}
\end{figure}

\para{The candidate check on eight models.} The perception check of \cref{sec:see} asks whether a model accepts a correct candidate more often than a wrong one, and whether this difference comes from the video. \cref{fig:cand} and \cref{tab:cand} extend it from the five models of the main paper to eight. With the video, every model accepts the correct candidate in at least 94.5\% of items and separates it from the wrong candidate by 51 to 97 points. Without the video, the six open-source models accept both candidates at nearly the same rate, with a difference of at most 10.5 points. The separation therefore comes from what the models see, and it is present in every model family that we test.

\begin{table}[t]
\centering
\scriptsize
\setlength{\tabcolsep}{3.6pt}
\begin{tabular}{@{}l ccc ccc@{}}
\toprule
 & \multicolumn{3}{c}{Video present} & \multicolumn{3}{c}{Video removed} \\
\cmidrule(lr){2-4}\cmidrule(l){5-7}
Model & Correct & Wrong & $\Delta$ & Correct & Wrong & $\Delta$ \\
\midrule
GPT-6.1 Terra & 98.5 & 2.0 & 96.5 & -- & -- & -- \\
Gemini 3.8 Flash & 97.5 & 1.0 & 96.5 & -- & -- & -- \\
Qwen3-VL-8B & 94.5 & 5.5 & 89.0 & 50.0 & 39.5 & 10.5 \\
Qwen3-VL-2B & 95.5 & 31.0 & 64.5 & 82.0 & 77.5 & 4.5 \\
InternVL3.5-8B & 98.0 & 29.0 & 69.0 & 70.0 & 64.5 & 5.5 \\
InternVideo2.5-8B & 95.0 & 20.0 & 75.0 & 50.5 & 45.0 & 5.5 \\
VideoLLaMA3-7B & 97.0 & 29.6 & 67.4 & 67.5 & 59.0 & 8.5 \\
LLaVA-Video-7B & 98.5 & 47.5 & 51.0 & 79.5 & 79.5 & 0.0 \\
\bottomrule
\end{tabular}
\caption{\textbf{Acceptance of embedded candidates (\%)} on the 200 Answerable base questions under the plain prompt. $\Delta$: correct minus wrong, in points. The proprietary models are evaluated with the video present.}
\label{tab:cand}
\end{table}

\para{Removing the video under the three-way prompt.} When the three-way prompt is used and no video is given, Qwen3-VL-2B and VideoLLaMA3-7B refuse nearly every question, including Answerable base questions, which shows that they notice the missing input once refusal is licensed. Under the plain prompt the same models answer 99\% of these questions, and with two candidates offered they pick the correct one at close to the chance level of 50\%.

\para{False premises.} Under the plain prompt, the three open-source models continue from a presupposed event in 19.1--37.5\% of cases and never refuse. The three-way prompt raises refusal to at most 26.0\% and, for Qwen3-VL-2B, raises acceptance to 71.0\%: the instruction licenses refusal, and the premise still goes unchecked.

\begin{table}[t]
\centering
\scriptsize
\setlength{\tabcolsep}{3.0pt}
\begin{tabular}{@{}l ccccc ccccc@{}}
\toprule
 & \multicolumn{5}{c}{Answerable: A-Acc} & \multicolumn{5}{c}{Guessable: G-Hit} \\
\cmidrule(lr){2-6}\cmidrule(l){7-11}
Model & +0 & +1 & +2 & +3 & +4 & +0 & +1 & +2 & +3 & +4 \\
\midrule
Qwen3-VL-2B & 90.1 & 91.1 & \underline{91.1} & \underline{91.1} & 91.1 & 75.0 & \underline{77.3} & 78.4 & 77.3 & 79.4 \\
Qwen3-VL-8B & \textbf{94.1} & \textbf{93.1} & \textbf{94.1} & \textbf{94.1} & \textbf{94.1} & \underline{78.4} & \underline{77.3} & \textbf{83.5} & \textbf{83.5} & \textbf{85.6} \\
VideoLLaMA3-7B & \underline{92.1} & \underline{92.1} & \textbf{94.1} & \textbf{94.1} & \underline{93.1} & \textbf{81.4} & \textbf{82.5} & \underline{79.4} & \underline{81.4} & \underline{83.5} \\
\bottomrule
\end{tabular}
\caption{\textbf{Redundant qualifiers} (plain prompt, \%): 101 Answerable and 97 Guessable base questions, each asked with zero to four added details that are true but not needed to identify the target. At most one question per level is refused.}
\label{tab:ladder}
\end{table}

\para{Redundant qualifiers.} Adding up to four needless details leaves every rate within a few points and never produces a guess (\cref{tab:ladder}). Together with the visibility claim in \cref{tab:induce}, this shows that the models are robust to wording that neither supplies a candidate answer nor changes the instruction.

\para{A reasoning model.} With explicit reasoning, Qwen3-VL-8B-Thinking refuses 89.2\% of blank-video questions (194 judged of 200) and 64.2\% of swapped-video questions (190 of 197) under the plain prompt, far more than its non-reasoning counterpart (2.5\% and 20.3\%), and it separates correct from wrong candidates by 88 points with the video. Under the three-way prompt it refuses all 200 blank-video questions and 92.4\% of the swapped ones. Reasoning therefore helps the model notice when the evidence is absent altogether. The graded cases remain: on the natural Unanswerable questions of the main set the same model refuses only 26.5\% under the plain prompt, and its G-Appr is 1.2\%.

\section{Fine-Tuning Details}
\label{supp:ft}

\begin{figure*}[t]
\centering
\scriptsize
\setlength{\tabcolsep}{4pt}
\renewcommand{\arraystretch}{1.2}
\begin{tabular}{@{}m{0.295\textwidth}lp{0.19\textwidth}p{0.125\textwidth}p{0.225\textwidth}@{}}
\toprule
\textbf{Training video} (Charades) & \textbf{Level} & \textbf{Question} & \textbf{Refusal-aware target} & \textbf{Guess-aware target} \\
\midrule
\multirow{3}{*}[-4pt]{\includegraphics[width=0.095\textwidth]{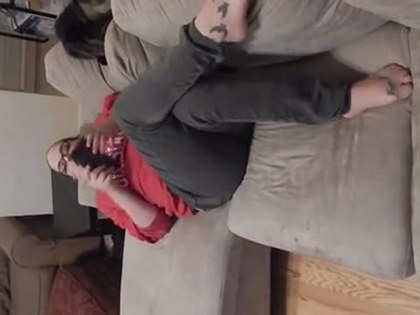}\,\includegraphics[width=0.095\textwidth]{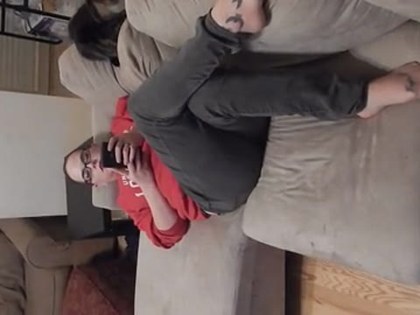}\,\includegraphics[width=0.095\textwidth]{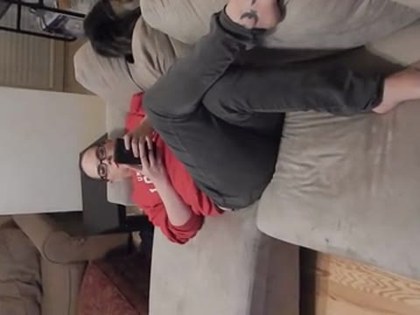}} & \lvA & Where does the woman take the phone in her hand from? & From the front pocket of her jeans. & From the front pocket of her jeans. \\
 & \lvG & What will the woman do next? & \cellcolor{cUl}Cannot be determined from the video. & \cellcolor{cGl}Probably keep lying on the sofa and looking at her phone, because after taking it out she stays half lying with her legs crossed, holds the phone in both hands, and does not change her posture until the end. \\
 & \lvU & What was the woman doing before she lay down on the sofa? & Cannot be determined from the video. & Cannot be determined from the video. \\
\bottomrule
\end{tabular}
\caption{\textbf{The two sets of training targets on one training video.} Each video carries questions of different levels, so the three targets appear side by side on the same footage. The two recipes share every input and every Answerable and Unanswerable target; they differ only in the highlighted cell, the target of the Guessable question. Guess-aware targets follow the form ``Probably X, because C'' with the reference guess X and its clue C.}
\label{fig:targets}
\end{figure*}

\begin{figure*}[t]
\centering
\footnotesize
\setlength{\tabcolsep}{3.5pt}
\renewcommand{\arraystretch}{1.15}
\begin{tabular}{@{}m{0.105\textwidth}p{0.155\textwidth}p{0.125\textwidth}p{0.115\textwidth}p{0.20\textwidth}p{0.20\textwidth}@{}}
\toprule
 & \textbf{Question} & \textbf{Base} & \textbf{Refusal-aware} & \textbf{Guess-aware} & \textbf{Guess-aware}$^\ddagger$ \\
\midrule
\includegraphics[width=\linewidth]{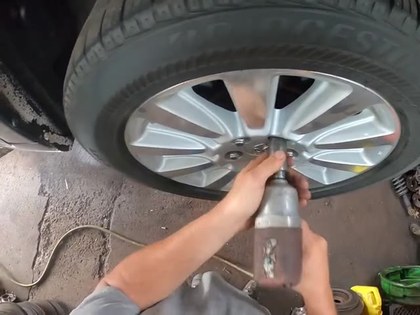} & \lvA\newline What type of tool does the wearer use to tighten the wheel nuts? & \tyA\ ``A pneumatic wrench.'' \yes & \tyA\ ``A pneumatic wrench.'' \yes & \tyA\ ``A pneumatic wrench.'' \yes & \tyA\ ``A pneumatic wrench.'' \yes \\
\includegraphics[width=\linewidth]{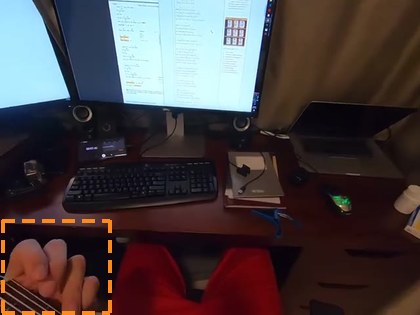} & \lvG\ \stag{SP}\newline What type of instrument is the object in the wearer's hand, of which only a small part is in view? & \tyR\ ``Cannot be determined from the video.'' \no & \tyR\ ``Cannot be determined from the video.'' \no & \tyG\ ``Probably a guitar or a ukulele, because the wearer's fingers press a row of thin metal strings at the bottom of the frame [\dots]'' \yes & \tyG\ ``Probably a guitar or a ukulele, because it is a string instrument with a long, thin neck [\dots] while its body is hidden by the wearer's hand and the desk.'' \yes \\
\includegraphics[width=\linewidth]{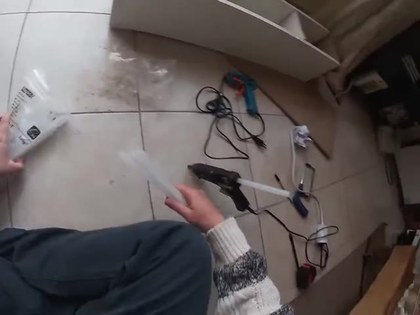} & \lvG\ \stag{TP}\newline After taking the white glue sticks out of the plastic bag, what will the wearer do next? & \tyA\ ``The person is going to use the white glue stick to glue something.'' \no & \tyR\ ``Cannot be determined from the video.'' \no & \tyG\ ``Probably load the glue stick into the hot glue gun, because the wearer took the glue gun from the shelf, plugged it in, and then took several white glue sticks out of the bag.'' \yes & \tyG\ ``Probably push the glue stick into the hot glue gun [\dots], because the wearer took this bag from the storage shelf and pulled the glue sticks out of it [\dots]'' \yes \\
\includegraphics[width=\linewidth]{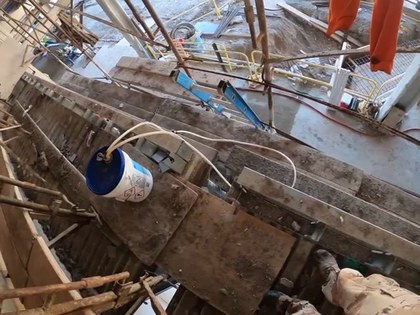} & \lvU\ \stag{PI}\newline What product name is printed on the label of the white bucket with the blue lid? & \tyR\ ``Cannot be determined from the video.'' \yes & \tyR\ ``Cannot be determined from the video.'' \yes & \tyR\ ``Cannot be determined from the video.'' \yes & \tyR\ ``Cannot be determined from the video.'' \yes \\
\bottomrule
\end{tabular}
\caption{\textbf{The same four questions before and after fine-tuning} (Qwen3-VL-8B, three-way prompt; all four clips are from Ego4D, which contributes no training video). \yes/\no: the response type matches or misses the expected one. $\ddagger$: trained with both prompts.}
\label{fig:beforeafter}
\end{figure*}

\para{Data.} The fine-tuning set has 608 questions on 228 Charades videos: 228 Answerable, 165 Guessable (44 spatial, 64 perceptual, 57 temporal), and 215 Unanswerable (75, 70, 70); 73 questions ask for a cause. Each video carries two or three questions of different levels (\cref{fig:targets}). Charades records which actor filmed each clip; all training videos come from actors who filmed none of the 299 benchmark clips, which also separates the homes. The split is 546 training questions (205 Answerable, 147 Guessable, 194 Unanswerable; 205 videos) and 62 validation questions (23, 18, 21; 23 videos).

\begin{table}[t]
\centering
\footnotesize
\setlength{\tabcolsep}{4pt}
\begin{tabular}{@{}ll@{}}
\toprule
Setting & Value \\
\midrule
Base model & Qwen3-VL-8B-Instruct \\
Trained parameters & LoRA on the language model \\
Vision encoder & frozen \\
LoRA rank / $\alpha$ / dropout & 16 / 32 / 0 \\
Optimizer & AdamW, weight decay 0.01 \\
Learning rate & $10^{-4}$, constant \\
Batch & 1 example, 8 accumulation steps \\
Gradient clipping & 1.0 \\
Precision & bfloat16 \\
Epochs / updates & 2 / 138 (274 with both prompts) \\
Random seed & 20261102 \\
Decoding at evaluation & greedy, 128 new tokens \\
\bottomrule
\end{tabular}
\caption{\textbf{Fine-tuning settings}, shared by all runs. The runs differ only in their targets and, for the both-prompts runs, in the prompts paired with each example; the runs of \cref{supp:second} replace the base model with Qwen3-VL-2B-Instruct.}
\label{tab:fthp}
\end{table}

\para{Training.} \cref{tab:fthp} lists the settings. Inputs use the three-way prompt and the frame sampling of evaluation. Training targets are written in the language of the annotations. The \emph{both prompts} runs pair every example with the plain and the three-way prompt (1,092 examples per epoch). For the refusal-aware run with both prompts, the sentence that permits a guess is removed from the three-way prompt during training and evaluation, so that the prompt and the targets agree; this row is marked in \cref{tab:ftfull}. The base row in all fine-tuning tables comes from the evaluation run of the fine-tuning pipeline, which differs slightly from the run in \cref{tab:main} (\eg A-Acc 90.3\% versus 89.3\% under the three-way prompt). In the fine-tuning tables, rates are computed over all questions, and the few invalid outputs count as failures.

\begin{table*}[t]
\centering
\scriptsize
\setlength{\tabcolsep}{5.2pt}
\begin{tabular}{@{}l ccc ccc ccc@{}}
\toprule
 & \multicolumn{3}{c}{Answerable} & \multicolumn{3}{c}{Guessable} & \multicolumn{3}{c}{Unanswerable} \\
\cmidrule(lr){2-4}\cmidrule(lr){5-7}\cmidrule(l){8-10}
Training & A-Acc & Guess & Refuse & Answer & G-Appr & Refuse & Answer & Guess & U-Ref \\
\midrule
\multicolumn{10}{@{}l}{\textit{Three-way prompt}}\\
Base & \textbf{90.3} & 0.0 & 2.7 & 47.1 & 1.4 & 50.3 & 12.5 & 0.8 & 86.7 \\
Refusal-aware & 79.9 & 0.0 & 15.7 & 8.4 & 0.0 & 91.6 & \textbf{0.8} & 0.0 & \textbf{99.2} \\
Refusal-aware, both prompts$^{\ast}$ & 82.8 & 0.0 & 11.0 & 10.7 & 0.0 & 89.3 & \underline{1.0} & 0.0 & \underline{99.0} \\
Guess-aware & 84.4 & 7.3 & \underline{2.1} & 21.2 & \underline{40.4} & \underline{30.8} & 3.5 & 9.2 & 87.3 \\
Guess-aware, both prompts & \underline{88.0} & 5.0 & \textbf{1.1} & 19.3 & \textbf{50.2} & \textbf{17.5} & 4.0 & 15.8 & 80.2 \\
\midrule
\multicolumn{10}{@{}l}{\textit{Plain prompt}}\\
Base & 89.1 & 0.0 & \textbf{0.0} & 98.4 & 0.2 & \textbf{0.2} & 85.7 & 2.1 & 10.6 \\
Refusal-aware & \underline{92.1} & 0.2 & \underline{0.3} & 66.0 & 2.1 & 31.7 & 25.5 & 0.3 & 74.2 \\
Refusal-aware, both prompts$^{\ast}$ & 85.9 & 0.0 & 8.2 & 11.7 & 0.0 & 88.3 & \textbf{1.4} & 0.0 & \textbf{98.6} \\
Guess-aware & \textbf{93.2} & 0.2 & \textbf{0.0} & 81.4 & \underline{11.7} & \underline{4.2} & 38.0 & 14.3 & 47.4 \\
Guess-aware, both prompts & 85.9 & 5.7 & 2.0 & 19.6 & \textbf{37.9} & 33.1 & \underline{2.9} & 8.4 & \underline{88.7} \\
\bottomrule
\end{tabular}
\caption{\textbf{Fine-tuning results under both evaluation prompts (\%).} $^{\ast}$Trained and evaluated with a variant of the three-way prompt that omits the sentence permitting a guess.}
\label{tab:ftfull}
\end{table*}

\begin{table*}[t]
\centering
\scriptsize
\setlength{\tabcolsep}{5.2pt}
\begin{tabular}{@{}l ccccc ccccc@{}}
\toprule
 & \multicolumn{5}{c}{Charades (source of the training videos)} & \multicolumn{5}{c}{Ego4D (unseen source)} \\
\cmidrule(lr){2-6}\cmidrule(l){7-11}
Training & A-Acc & A-FR & G-Appr & G-FR & U-Ref & A-Acc & A-FR & G-Appr & G-FR & U-Ref \\
\midrule
Base & \textbf{90.5} & 2.9 & 0.0 & 62.9 & 89.0 & \textbf{89.5} & 2.1 & 1.6 & 48.2 & \underline{84.4} \\
Refusal-aware & 80.5 & 15.1 & 0.0 & 98.4 & \textbf{99.4} & 77.6 & 18.2 & 0.0 & 90.5 & \textbf{99.1} \\
Refusal-aware, both prompts$^{\ast}$ & 84.0 & 10.2 & 0.0 & 95.2 & \underline{99.0} & 78.3 & 14.0 & 0.0 & 88.3 & \textbf{99.1} \\
Guess-aware & 84.2 & \underline{2.3} & \underline{33.9} & \underline{38.7} & 93.2 & 85.3 & \underline{1.4} & \underline{41.5} & \underline{29.4} & 81.6 \\
Guess-aware, both prompts & \underline{88.4} & \textbf{1.2} & \textbf{45.2} & \textbf{22.6} & 84.5 & \underline{86.7} & \textbf{0.7} & \textbf{51.1} & \textbf{16.6} & 76.0 \\
\bottomrule
\end{tabular}
\caption{\textbf{Fine-tuning results by video source} (three-way prompt, \%). Under guess-aware tuning, G-Appr on Ego4D rises from 1.6\% to 41.5\% (51.1\% with both prompts), a gain at least as large as on Charades, although Ego4D contributes no training video.}
\label{tab:ftsrc}
\end{table*}

\para{Results under the plain prompt.} \cref{tab:ftfull} shows that single-prompt tuning transfers partly to the plain prompt: the refusal-aware model refuses 74.2\% of Unanswerable questions there, and the guess-aware model 47.4\%, with a G-Appr of 11.7\%. With both prompts in training, the guess-aware model behaves similarly under either prompt, so the earlier dependence on the prompt comes from the training protocol.

\para{Statistical tests.} All comparisons with the base model are paired by question, with 2,000 bootstrap resamples over videos and invalid outputs counted as failures; the intervals are in \cref{tab:ft}. On Ego4D alone, guess-aware tuning changes G-Appr by $+39.9$ [35.2, 44.9], and guess-aware tuning with both prompts by $+49.5$ [44.5, 54.3].

\para{Side effects.} The perturbation subset used after fine-tuning has 600 items under the three-way prompt: 200 two-candidate questions, 200 blank-video questions, and 100 Unanswerable questions each with a wrong candidate and with a visibility claim. Two-candidate accuracy is 97.5\% for the base model, 91.0\% after guess-aware tuning, and 95.0\% with both prompts; refusal under a visibility claim is 79.0\%, 86.0\%, and 80.0\%. Blank-video refusal is in \cref{tab:ft}; adoption of a wrong candidate is 28.0\%, 39.0\%, and 46.0\%. After tuning, most of this adoption is hedged: the share of wrong candidates stated as fact falls from 23\% for the base model to 5\% and 7\%, while 34\% and 39\% are adopted as a guess, so the tuned models treat the candidate in the question as a clue. The training set contains neither blank videos nor embedded candidates, so these two conditions lie outside the supervision, and adding such examples to the training set is the direct way to cover them. Video-MME is evaluated on a fixed subset of 500 questions from 168 videos with the official answer parser; the refusal-aware models were not evaluated on these two sets.

\para{Examples.} \cref{fig:beforeafter} follows four benchmark questions through fine-tuning. The two Guessable rows show the two ways in which the base model misses the middle response, refusal and assertion, and both are replaced by a guess that cites its clue. The response format ``Probably X, because C'' matters for the user: in both rows the stated clue (the strings under the fingers, the glue gun that was just plugged in) can be verified in the frames. The first and last rows show that the responses to full and to absent evidence stay in place.

\section{Fine-Tuning a Second Base Model}
\label{supp:second}

\textbf{The contrast between the two sets of targets holds on a second base model.} We repeat the experiment of Sec.~5 on Qwen3-VL-2B with the same data, targets, and settings (\cref{tab:fthp}). The 2B model is a useful second case because it starts from the opposite side: under the three-way prompt the 8B base model already refuses 86.7\% of Unanswerable questions, whereas the 2B base model refuses only 40.9\% and answers most questions at every level.

\begin{figure*}[t]
\centering
\includegraphics[width=\textwidth]{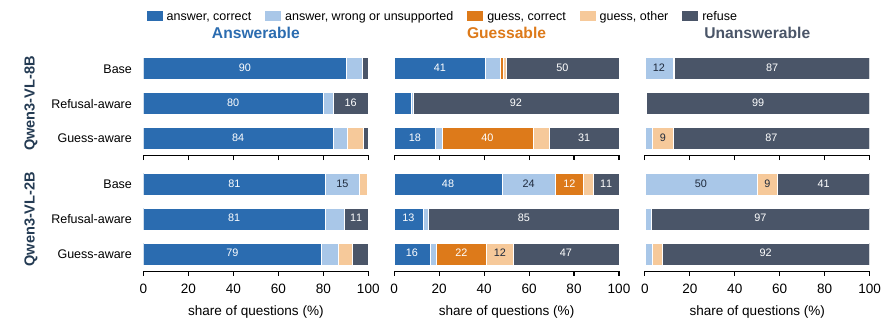}
\caption{\textbf{Response types before and after fine-tuning on two base models} (three-way prompt). On both models, refusal-aware targets turn the Guessable row gray, and guess-aware targets open the dark orange band of grounded guesses there while the other two levels keep their expected response.}
\label{fig:ft2b}
\end{figure*}

\begin{table}[t]
\centering
\scriptsize
\setlength{\tabcolsep}{3.2pt}
\begin{tabular}{@{}l ccccc cc@{}}
\toprule
Training & A-Acc & A-FR $\downarrow$ & G-Appr & G-FR $\downarrow$ & U-Ref & Avg & ES \\
\midrule
\multicolumn{8}{@{}l}{\emph{Qwen3-VL-2B}} \\
Base & \textbf{80.9} & \textbf{0.6} & \underline{12.4} & \textbf{11.4} & 40.9 & 44.7 & 13.5 \\
Refusal-aware & \underline{80.6} & 10.7 & 0.0 & 84.6 & \textbf{97.1} & \underline{59.3} & \underline{35.5} \\
Guess-aware & 79.1 & \underline{7.1} & \textbf{22.5} & \underline{47.1} & \underline{92.4} & \textbf{64.7} & \textbf{38.7} \\
\midrule
\multicolumn{8}{@{}l}{\emph{Qwen3-VL-8B (Tab.~2 and Tab.~4 of the main paper)}} \\
Base & \textbf{90.3} & \underline{2.7} & \underline{1.4} & \underline{50.3} & 86.7 & 59.5 & 31.0 \\
Refusal-aware & 79.9 & 15.7 & 0.0 & 91.6 & \textbf{99.2} & \underline{59.7} & \underline{36.1} \\
Guess-aware & \underline{84.4} & \textbf{2.1} & \textbf{40.4} & \textbf{30.8} & \underline{87.3} & \textbf{70.7} & \textbf{43.9} \\
\bottomrule
\end{tabular}
\caption{\textbf{Fine-tuning on two base models} (three-way prompt, \%). The base rows are evaluated in the fine-tuning pipeline.}
\label{tab:second}
\end{table}

\para{Both hypotheses are reproduced.} \cref{fig:ft2b} and \cref{tab:second} show the same picture as on the 8B model. Refusal-aware targets remove guessing altogether (G-Appr from 12.4\% to 0.0\%) and refuse 84.6\% of Guessable questions; false refusals on Answerable questions rise from 0.6\% to 10.7\%, a paired increase of 10.0 points [7.7, 12.3]. Guess-aware targets raise G-Appr from 12.4\% to 22.5\%, a paired gain of 10.1 points [5.4, 14.8], and reach the highest Avg (64.7\%) and ES (38.7) of the three models. The guesses again land where the clues are: 34.2\% of Guessable questions receive a guess, against 6.1\% of Answerable and 4.4\% of Unanswerable ones.

\para{Guess-aware targets give the better trade-off at every level.} Compared with refusal-aware tuning on the same 2B model, guess-aware tuning refuses 37.5 points fewer Guessable questions (47.1\% versus 84.6\%) and fewer Answerable ones (7.1\% versus 10.7\%), while U-Ref stays above 92\%. The learned response also carries over to the unseen source: G-Appr on Ego4D rises from 13.4\% to 21.4\%.

\para{What the smaller model has to learn.} The two base models differ in how much the fine-tuning has to change. The 8B base model already refuses most Unanswerable questions, so guess-aware tuning only has to move part of its refusals on Guessable questions to guesses, and its A-FR stays at 2.1\%. The 2B base model rarely refuses, so either recipe must first establish refusal itself: U-Ref rises by more than 50 points under both. Part of this newly learned refusal reaches Answerable questions, and it does so under both recipes (A-FR of 7.1\% and 10.7\%). The 2B model thus pays for acquiring refusal, and it pays less when partial evidence has its own target. For a model that has no working refusal yet, the two skills are learned together, and the guess level keeps the new refusal from absorbing the questions that carry clues.

\section{Questions from an External Benchmark}
\label{supp:star}

\textbf{The learned guess carries over to questions that we did not write.} All Guessable questions in the main paper come from our own annotation. To test the fine-tuned models on questions written by others, we use the Prediction questions of STAR~[\textcolor{cvprblue}{S1}], which show a clip of a person at home and ask what the person will do next. The answer lies after the clip and the visible activity narrows it, so these questions correspond to our Guessable level with a temporal gap. We sample 300 questions from the validation set and format them to match our evaluation: each question is asked open-ended under the three-way prompt, without its answer options, and in the language of our annotations. The judge compares the content with the text of the correct option.

\begin{figure}[t]
\centering
\includegraphics[width=\linewidth]{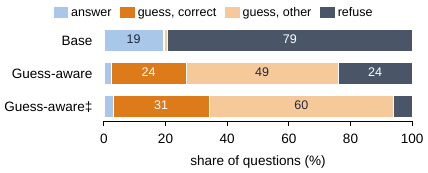}
\caption{\textbf{Responses to 300 STAR Prediction questions} (three-way prompt). The base model refuses most of them, and the fine-tuned models answer with guesses. $\ddagger$: trained with both prompts.}
\label{fig:star}
\end{figure}

\begin{table}[t]
\centering
\footnotesize
\setlength{\tabcolsep}{5pt}
\begin{tabular}{@{}l cccc@{}}
\toprule
Qwen3-VL-8B & Guess & Correct guess & Refuse $\downarrow$ & Hit \\
\midrule
Base & 1.3 & 0.3 & 79.3 & 9.7 \\
+ Guess-aware & \underline{73.7} & \underline{24.3} & \underline{24.0} & \underline{25.7} \\
+ Guess-aware$^\ddagger$ & \textbf{91.0} & \textbf{31.1} & \textbf{6.0} & \textbf{33.4} \\
\bottomrule
\end{tabular}
\caption{\textbf{STAR Prediction questions (\%).} Guess: responses of the Guess type. Correct guess: guesses whose content matches the reference. Hit: correct content under an answer or a guess.}
\label{tab:star}
\end{table}

\para{From refusal to grounded guesses.} The base model treats these questions as it treats our Guessable questions: it refuses 79.3\% of them and guesses on 1.3\% (\cref{fig:star}, \cref{tab:star}). The guess-aware model guesses on 73.7\%, and its correct guesses rise from 0.3\% to 24.3\%. Its hit rate (25.7\%) is more than twice that of the base model (9.7\%), so the guesses add correct content that the base model withheld. Predicting the next action without answer options is a hard task, and about one guess in three names the action that follows. The model was trained on 147 guess targets from our own annotation, and neither the STAR questions nor their source annotations appear in training; none of the sampled clips comes from a training video.

\para{Training with both prompts transfers further.} The model trained with both prompts guesses on 91.0\% of the questions and reaches 31.1\% correct guesses with a hit rate of 33.4\%, while refusing only 6.0\%. This matches Sec.~5.3 of the main paper, where the same model gave the highest G-Appr on our own questions: more varied training inputs make the guess response less dependent on the exact form of the input.

\section{Few-Shot Demonstrations as a Baseline}
\label{supp:fewshot}

\textbf{Showing the response format in the prompt recovers a small part of what supervision provides.} The three-way prompt already describes the guess response in words. A natural next step without training is to demonstrate it. We add three text demonstrations from other videos to the three-way prompt, one per level, taken from the fine-tuning set with their guess-aware targets, so the prompt contains a complete ``Probably X, because C'' example.

\begin{promptbox}[Three-way prompt with demonstrations]
\emph{(the four instruction lines of the three-way prompt)}\\
Examples from other videos:\\
Question: \{Answerable question\}\quad Answer: \{answer\}\\
Question: \{Guessable question\}\quad Answer: Probably \{X\}, because \{C\}.\\
Question: \{Unanswerable question\}\quad Answer: Cannot be determined from the video.\\
Question: \{question\}
\end{promptbox}

\begin{table}[t]
\centering
\scriptsize
\setlength{\tabcolsep}{3.4pt}
\begin{tabular}{@{}l ccccc cc@{}}
\toprule
Qwen3-VL-8B & A-Acc & A-FR $\downarrow$ & G-Appr & G-FR $\downarrow$ & U-Ref & Avg & ES \\
\midrule
Base & \textbf{90.3} & 2.7 & 1.4 & 50.3 & \underline{86.7} & 59.5 & 31.0 \\
+ 3 demonstrations & \underline{89.9} & \underline{1.7} & 9.3 & 34.5 & 79.1 & 59.4 & 30.1 \\
+ Guess-aware & 84.4 & 2.1 & \underline{40.4} & \underline{30.8} & \textbf{87.3} & \underline{70.7} & \underline{43.9} \\
+ Guess-aware$^\ddagger$ & 88.0 & \textbf{1.1} & \textbf{50.2} & \textbf{17.5} & 80.2 & \textbf{72.8} & \textbf{48.5} \\
\bottomrule
\end{tabular}
\caption{\textbf{Demonstrations versus fine-tuning} (three-way prompt, \%). The first two rows are evaluated in the fine-tuning pipeline; the last two repeat Tab.~2 of the main paper. $\ddagger$: trained with both prompts.}
\label{tab:fewshot}
\end{table}

\para{Results.} With the demonstrations, G-Appr rises from 1.4\% to 9.3\%, a paired gain of 7.9 points [5.3, 10.7], and refusals of Guessable questions fall from 50.3\% to 34.5\% (\cref{tab:fewshot}). The demonstrations make the model more willing to give content at every level, however: U-Ref falls by 7.6 points [5.5, 9.9], so Avg (59.4\%) and ES (30.1) stay at the level of the base model (59.5\% and 31.0). The model shifts its threshold for speaking and does not become better at telling partial from absent evidence. Guess-aware fine-tuning reaches a G-Appr of 40.4\% with U-Ref unchanged and raises ES to 43.9. It also needs no demonstrations at test time: the both-prompts model guesses appropriately on 37.9\% of Guessable questions under the plain prompt, which mentions neither guessing nor refusal.

\section{What a Response Type Tells the User}
\label{supp:calib}

\textbf{A hedge marks lower certainty, and fine-tuning makes each response type more reliable.} The main paper argues that a response type should tell the user how much evidence lies behind the content. A direct test is to compare how often the content is correct under each type. For each model under the three-way prompt we compute the accuracy of its definite answers and of its guesses on Answerable and Guessable questions, and the share of its refusals that fall on Unanswerable questions.

\begin{figure}[t]
\centering
\includegraphics[width=\linewidth]{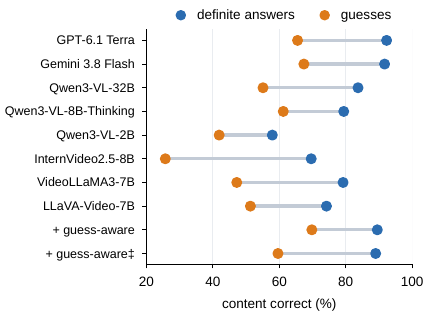}
\caption{\textbf{Accuracy of definite answers and of guesses} (three-way prompt; models with at least 20 guesses). $\ddagger$: trained with both prompts.}
\label{fig:calib}
\end{figure}

\begin{table}[t]
\centering
\scriptsize
\setlength{\tabcolsep}{3pt}
\begin{tabular}{@{}l rc rc rc@{}}
\toprule
 & \multicolumn{2}{c}{Answers} & \multicolumn{2}{c}{Guesses} & \multicolumn{2}{c}{Refusals} \\
\cmidrule(lr){2-3}\cmidrule(lr){4-5}\cmidrule(l){6-7}
Model & \# & correct & \# & correct & \# & on U \\
\midrule
GPT-6.1 Terra & 766 & 92.3 & 377 & 65.5 & 578 & 85.1 \\
Gemini 3.8 Flash & 840 & 91.7 & 168 & 67.4 & 682 & 78.1 \\
Qwen3-VL-32B & 926 & 83.7 & 197 & 55.1 & 599 & 82.0 \\
Qwen3-VL-8B & 920 & 82.3 & 15 & -- & 786 & 69.7 \\
\midrule
+ Refusal-aware & 598 & \textbf{93.8} & 0 & -- & 1,123 & 55.7 \\
+ Guess-aware & 712 & 89.5 & 311 & \textbf{69.8} & 697 & 79.1 \\
+ Guess-aware$^\ddagger$ & 729 & 89.0 & 403 & 59.6 & 588 & \textbf{86.1} \\
\bottomrule
\end{tabular}
\caption{\textbf{Reliability of each response type} (three-way prompt). Correct: share of responses of this type with correct content, in \%. On U: share of refusals that fall on Unanswerable questions. The lower block fine-tunes Qwen3-VL-8B. Guess accuracy is omitted below 20 guesses.}
\label{tab:calib}
\end{table}

\para{Answers are more accurate than guesses for every model.} \cref{fig:calib} shows that all models that guess at least 20 times are more often right when they answer than when they hedge, by 16 to 44 points. When a model says ``probably'', its content is in fact less certain, so the three response types are informative and worth scoring separately.

\para{Fine-tuning sharpens all three types.} After guess-aware tuning, the guesses of Qwen3-VL-8B are correct in 69.8\% of cases, the highest value of all models (\cref{tab:calib}), and its definite answers become more accurate as well (82.3\% to 89.5\%), as uncertain content moves from the answer type to the guess type. Its refusals also land on Unanswerable questions more often (69.7\% to 79.1\%, and 86.1\% with both prompts). Refusal-aware tuning moves in the opposite direction on this last measure: only 55.7\% of its 1,123 refusals fall on Unanswerable questions, since it also refuses the Guessable and many Answerable ones. A user of the guess-aware model can therefore read more from each response type than a user of either the base or the refusal-aware model.

\section{Qualitative Examples}
\label{supp:qual}
\newcommand{\wrow}[3]{#1 & #2 & #3 \\}
\begin{figure*}[!t]
\centering
\footnotesize
\setlength{\tabcolsep}{3pt}
\renewcommand{\arraystretch}{1.12}
\begin{minipage}[t]{0.165\textwidth}\vspace{0pt}
\begin{tikzpicture}[inner sep=0]\node(a){\includegraphics[width=\linewidth]{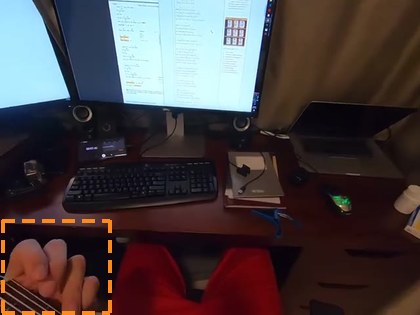}};\node[anchor=south west,fill=black,fill opacity=.55,text opacity=1,text=white,font=\tiny\sffamily,inner sep=1.2pt] at (a.south west){9 s};\end{tikzpicture}\\[2pt]\begin{tikzpicture}[inner sep=0]\node(a){\includegraphics[width=\linewidth]{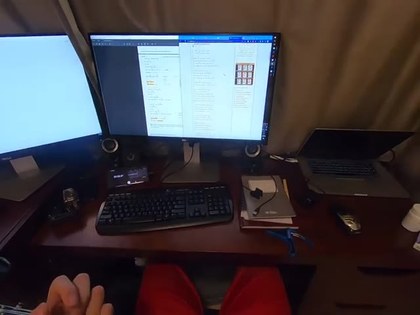}};\node[anchor=south west,fill=black,fill opacity=.55,text opacity=1,text=white,font=\tiny\sffamily,inner sep=1.2pt] at (a.south west){10 s};\end{tikzpicture}\\[3pt]
\lvG\ \stag{SP}\\[2pt]
\textbf{Q:} \emph{``What type of instrument is the object in the wearer's hand, of which only a small part is in view?''}\\[2pt]
\textbf{Reference guess:} a guitar; a bass; a ukulele.\\
\textbf{Clue:} the neck and strings at the lower left.
\end{minipage}\hfill
\begin{minipage}[t]{0.82\textwidth}\vspace{0pt}
\begin{tabular}{@{}lp{0.27\linewidth}p{0.51\linewidth}@{}}
\toprule
Model & Plain prompt & Three-way prompt \\
\midrule
\wrow{GPT-6.1 Terra}{\tyA\ ``A guitar.''}{\tyG\ ``It looks like a guitar.'' \yes}
\wrow{Gemini 3.8 Flash}{\tyG\ ``A guitar (or a ukulele; more likely a guitar).'' \yes}{\tyG\ ``At the lower left (about 0:08--0:10) the neck, frets, and strings are visible and the wearer's hand presses the strings [\dots] presumably a guitar (or a ukulele).'' \yes}
\wrow{Qwen3-VL-32B}{\tyA\ ``Guitar''}{\tyG\ ``[\dots] several metal strings are clearly visible [\dots] it is very likely a plucked string instrument such as a guitar or a ukulele.'' \yes}
\wrow{Qwen3-VL-8B-Thinking}{\tyA\ ``ukulele''}{\tyG\ ``[\dots] A guitar is a reasonable guess, but this cannot be confirmed with certainty from the video.'' \yes}
\wrow{Qwen3-VL-8B}{\tyA\ ``guitar''}{\tyR\ ``Cannot be determined from the video.''}
\wrow{Qwen3-VL-2B}{\tyA\ ``Guitar''}{\tyA\ ``The object in the wearer's hand [\dots] is a guitar.''}
\wrow{InternVL3.5-8B}{\tyA\ ``Guitar''}{\tyA\ ``A guitar.''}
\wrow{InternVideo2.5-8B}{\tyA\ ``A guitar.''}{\tyA\ ``A guitar.''}
\wrow{VideoLLaMA3-7B}{\tyA\ ``A guitar.''}{\tyR\ ``Cannot be determined from the video.''}
\wrow{LLaVA-Video-7B}{\tyA\ ``guitar.''}{\tyG\ ``The object in the person's hand is a stringed instrument, likely a guitar.'' \yes}
\bottomrule
\end{tabular}
\end{minipage}\\[10pt]
\begin{minipage}[t]{0.165\textwidth}\vspace{0pt}
\begin{tikzpicture}[inner sep=0]\node(a){\includegraphics[width=\linewidth]{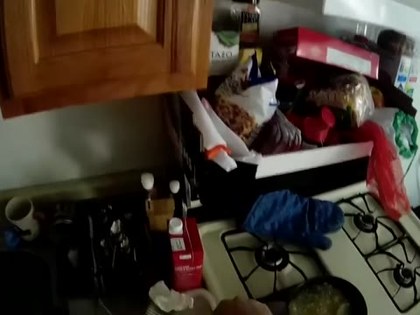}};\node[anchor=south west,fill=black,fill opacity=.55,text opacity=1,text=white,font=\tiny\sffamily,inner sep=1.2pt] at (a.south west){5 s};\end{tikzpicture}\\[2pt]\begin{tikzpicture}[inner sep=0]\node(a){\includegraphics[width=\linewidth]{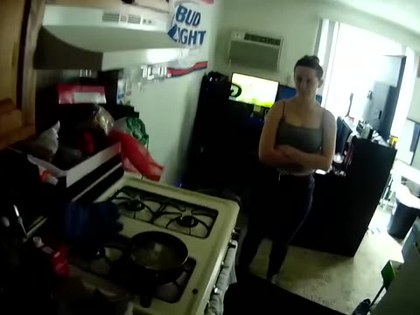}};\node[anchor=south west,fill=black,fill opacity=.55,text opacity=1,text=white,font=\tiny\sffamily,inner sep=1.2pt] at (a.south west){16 s};\end{tikzpicture}\\[3pt]
\lvU\ \stag{SP}\ \stag{CL}\\[2pt]
\textbf{Q:} \emph{``Why is the woman standing by the stove with her arms crossed, looking at the camera wearer?''}\\[2pt]
\textbf{Missing evidence:} what the wearer is doing off camera; the clip is silent.
\end{minipage}\hfill
\begin{minipage}[t]{0.82\textwidth}\vspace{0pt}
\begin{tabular}{@{}lp{0.36\linewidth}p{0.42\linewidth}@{}}
\toprule
Model & Plain prompt & Three-way prompt \\
\midrule
\wrow{GPT-6.1 Terra}{\tyR\ ``The video does not make the reason clear [\dots]'' \yes}{\tyR\ ``It cannot be determined from the video alone [\dots]'' \yes}
\wrow{Gemini 3.8 Flash}{\tyA\ ``Because she is waiting for the wearer to cook (or for the food to be ready).''}{\tyR\ ``Cannot be determined from the video.'' \yes}
\wrow{Qwen3-VL-32B}{\tyA\ ``She is watching the wearer cook.''}{\tyR\ ``Cannot be determined from the video.'' \yes}
\wrow{Qwen3-VL-8B-Thinking}{\tyA\ ``She is waiting for the person to finish cooking.''}{\tyG\ ``[\dots] It is plausible she is observing the person filming or waiting for an action [\dots]''}
\wrow{Qwen3-VL-8B}{\tyA\ ``She is watching the person cook.''}{\tyR\ ``Cannot be determined from the video.'' \yes}
\wrow{Qwen3-VL-2B}{\tyG\ ``She may be waiting for the wearer to finish some task.''}{\tyG\ ``She may be waiting for the wearer to finish some task, or observing what the wearer does.''}
\wrow{InternVL3.5-8B}{\tyA\ ``Waiting for the wearer to wash the dishes.''}{\tyA\ ``Waiting for the wearer to wash the dishes.''}
\wrow{InternVideo2.5-8B}{\tyA\ ``Supervising the wearer.''}{\tyA\ ``Waiting for the wearer to cook.''}
\wrow{VideoLLaMA3-7B}{\tyA\ ``waiting for food.''}{\tyR\ ``Cannot be determined from the video.'' \yes}
\wrow{LLaVA-Video-7B}{\tyA\ ``She is waiting for the food to cook.''}{\tyG\ ``[\dots] possibly indicating she is waiting or observing something.''}
\bottomrule
\end{tabular}
\end{minipage}
\caption{\textbf{Two questions, ten models, two prompts.} Top: a Guessable question whose clue is the neck and strings at the lower left of the frame; under the three-way prompt the same clue yields guesses, assertions, and refusals. Bottom: a causal Unanswerable question; nothing in the clip explains the woman's posture. \yes: the expected response type.}
\label{fig:wall}
\end{figure*}

\begin{figure*}[!t]
\centering
\footnotesize
\setlength{\tabcolsep}{3pt}
\renewcommand{\arraystretch}{1.12}
\begin{minipage}[t]{0.165\textwidth}\vspace{0pt}
\begin{tikzpicture}[inner sep=0]\node(a){\includegraphics[width=\linewidth]{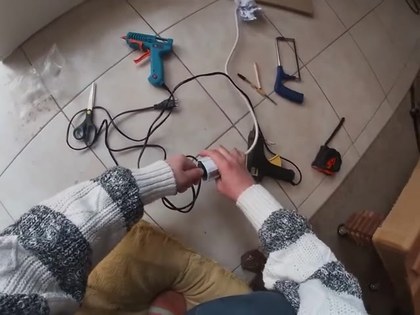}};\node[anchor=south west,fill=black,fill opacity=.55,text opacity=1,text=white,font=\tiny\sffamily,inner sep=1.2pt] at (a.south west){11 s};\end{tikzpicture}\\[2pt]\begin{tikzpicture}[inner sep=0]\node(a){\includegraphics[width=\linewidth]{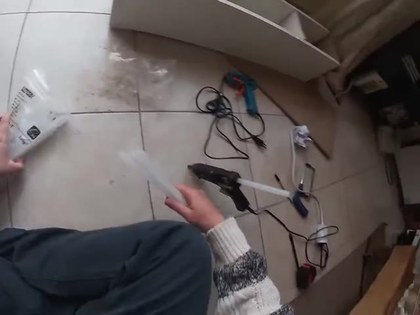}};\node[anchor=south west,fill=black,fill opacity=.55,text opacity=1,text=white,font=\tiny\sffamily,inner sep=1.2pt] at (a.south west){36 s};\end{tikzpicture}\\[3pt]
\lvG\ \stag{TP}\\[2pt]
\textbf{Q:} \emph{``After taking the white glue sticks out of the plastic bag, what will the wearer do next?''}\\[2pt]
\textbf{Reference guess:} load a stick into the hot glue gun; glue the twine onto the paper tube.\\
\textbf{Clue:} the plugged-in glue gun on the floor.
\end{minipage}\hfill
\begin{minipage}[t]{0.82\textwidth}\vspace{0pt}
\begin{tabular}{@{}lp{0.38\linewidth}p{0.40\linewidth}@{}}
\toprule
Model & Plain prompt & Three-way prompt \\
\midrule
\wrow{GPT-6.1 Terra}{\tyA\ ``Put the glue stick next to the hot glue gun, ready for use.''}{\tyA\ ``The wearer then puts the glue stick next to the hot glue gun, ready for use.''}
\wrow{Gemini 3.8 Flash}{\tyA\ ``After putting the glue sticks on the floor, the wearer reaches for the black hot glue gun.''}{\tyR\ ``Cannot be determined (the clip ends after the wearer places the glue sticks on the floor [\dots]).''}
\wrow{Qwen3-VL-32B}{\tyA\ ``Insert the glue stick into the hot glue gun.''}{\tyA\ ``[\dots] the wearer then inserts the glue stick into the hot glue gun, ready for use.''}
\wrow{Qwen3-VL-8B-Thinking}{\tyA\ ``Use the white glue stick with the hot glue gun.''}{\tyA\ ``placing the white glue stick on the floor''}
\wrow{Qwen3-VL-8B}{\tyA\ ``Put the glue stick on the floor.''}{\tyA\ ``The person is going to use the white glue stick to glue something.''}
\wrow{Qwen3-VL-2B}{\tyA\ ``Put the glue stick on the floor.''}{\tyA\ ``[\dots] the wearer is going to put the glue stick on the floor.''}
\wrow{InternVL3.5-8B}{\tyA\ ``connect the wire''}{\tyA\ ``Insert the glue stick into the hot glue gun.''}
\wrow{InternVideo2.5-8B}{\tyA\ ``Open the glue stick.''}{\tyA\ ``Install the glue stick.''}
\wrow{VideoLLaMA3-7B}{\tyA\ ``Insert the glue stick into the hot glue gun.''}{\tyA\ ``Insert the glue stick into the hot glue gun.''}
\wrow{LLaVA-Video-7B}{\tyA\ ``put the white glue stick into the hot glue gun.''}{\tyA\ ``Insert the white glue stick into the hot glue gun.''}
\bottomrule
\end{tabular}
\end{minipage}\\[10pt]
\begin{minipage}[t]{0.165\textwidth}\vspace{0pt}
\begin{tikzpicture}[inner sep=0]\node(a){\includegraphics[width=\linewidth]{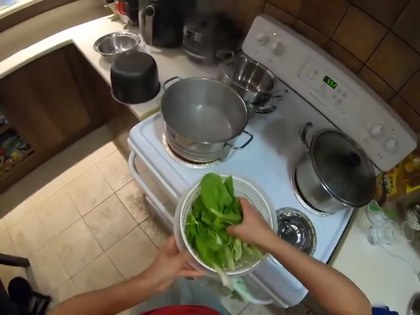}};\node[anchor=south west,fill=black,fill opacity=.55,text opacity=1,text=white,font=\tiny\sffamily,inner sep=1.2pt] at (a.south west){8 s};\end{tikzpicture}\\[2pt]\begin{tikzpicture}[inner sep=0]\node(a){\includegraphics[width=\linewidth]{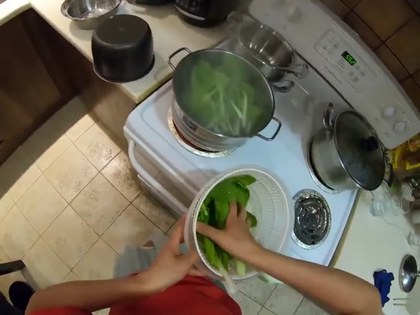}};\node[anchor=south west,fill=black,fill opacity=.55,text opacity=1,text=white,font=\tiny\sffamily,inner sep=1.2pt] at (a.south west){14 s};\end{tikzpicture}\\[3pt]
\lvU\ \stag{TP}\\[2pt]
\textbf{Q:} \emph{``For how long will the wearer boil these greens?''}\\[2pt]
\textbf{Missing evidence:} the cooking time lies after the clip.
\end{minipage}\hfill
\begin{minipage}[t]{0.82\textwidth}\vspace{0pt}
\begin{tabular}{@{}lp{0.38\linewidth}p{0.40\linewidth}@{}}
\toprule
Model & Plain prompt & Three-way prompt \\
\midrule
\wrow{GPT-6.1 Terra}{\tyR\ ``The cooking time cannot be determined from the video alone.'' \yes}{\tyR\ ``Cannot be determined from the video.'' \yes}
\wrow{Gemini 3.8 Flash}{\tyR\ ``The video does not tell (the cooking time is neither mentioned nor shown).'' \yes}{\tyR\ ``The video does not provide enough information [\dots]'' \yes}
\wrow{Qwen3-VL-32B}{\tyR\ ``I don't know.'' \yes}{\tyR\ ``Cannot be determined from the video.'' \yes}
\wrow{Qwen3-VL-8B-Thinking}{\tyR\ ``The video does not provide information about the cooking duration for the greens.'' \yes}{\tyR\ ``Cannot be determined from the video.'' \yes}
\wrow{Qwen3-VL-8B}{\tyA\ ``1 minute.''}{\tyR\ ``Cannot be determined from the video.'' \yes}
\wrow{Qwen3-VL-2B}{\tyA\ ``10 minutes.''}{\tyR\ ``[\dots] how long they are boiled is not shown in the video.'' \yes}
\wrow{InternVL3.5-8B}{\tyA\ ``1 minute.''}{\tyA\ ``1 minute.''}
\wrow{InternVideo2.5-8B}{\tyA\ ``Three minutes.''}{\tyR\ ``Cannot be determined from the video.'' \yes}
\wrow{VideoLLaMA3-7B}{\tyA\ ``Three minutes.''}{\tyR\ ``Cannot be determined from the video.'' \yes}
\wrow{LLaVA-Video-7B}{\tyA\ ``3 minutes.''}{\tyR\ ``Cannot be determined from the video.'' \yes}
\bottomrule
\end{tabular}
\end{minipage}
\caption{\textbf{Two temporal gaps.} Top: a Guessable question; the next step lies after the clip, and the glue gun on the floor hints at it. None of the twenty responses is a guess. Bottom: an Unanswerable question; nothing in the clip indicates a cooking time. \yes: the expected response type.}
\label{fig:wall2}
\end{figure*}

\cref{fig:wall,fig:wall2} show four questions with the responses of all ten models under both prompts. Three observations of Sec.~4 can be read off these cases.

\para{The clue is seen.} On both Guessable questions most models name an answer that the clue supports; for the guitar neck, all ten do so under the plain prompt.

\para{Its weight is judged differently by every model.} Under the three-way prompt, the guitar neck leads five models to hedge, three to assert, and two to refuse, and the two refusals withdraw an answer that was correct under the plain prompt. For the glue gun, the temporal gap goes unnoticed: nineteen of the twenty responses report the next step as if it had been observed.

\para{Without evidence, the content comes from commonsense.} Eight models explain the woman's posture under the plain prompt, with reasons that range from waiting for food to supervising the wearer, and six name a cooking time for the greens, between one and ten minutes. The reasons and the durations differ from model to model, with cooking times a factor of ten apart, which points to commonsense as their source.

\section{Release}
\label{supp:release}

To make all results reproducible, we will release the annotated questions with their levels, tags, reference guesses, and clues, the perturbation set, the fine-tuning set with both sets of targets, the prompts, the judge prompts, the scoring code, and all other materials needed to rerun the experiments at \url{https://github.com/suat-llm/ClueProbe}. Videos are referenced by their identifiers in Charades and Ego4D and obtained from the original providers. One record of the main set looks as follows.

\begin{promptbox}[Example record]
\ttfamily\scriptsize
video: Ego4D / ef797776-\dots\quad level: Guessable\quad gap: PI\quad causal: no\\
question: What type of tool is the thin red object in the wearer's right hand?\\
reference\_guess: a pencil\\
clue: the wearer holds it along a tape measure and marks the board with it
\end{promptbox}

\newpage
\section*{Additional Reference}
{\small\noindent[S1] Bo Wu, Shoubin Yu, Zhenfang Chen, Joshua~B. Tenenbaum, and Chuang Gan. STAR: A benchmark for situated reasoning in real-world videos. In \emph{NeurIPS Datasets and Benchmarks Track}, 2021.}

\end{document}